\documentclass{article}

\PassOptionsToPackage{numbers, compress}{natbib}
\usepackage[table]{xcolor}
\usepackage{array}

\usepackage[main, final]{neurips_2026}

\usepackage[utf8]{inputenc} 
\usepackage[T1]{fontenc}    
\usepackage{hyperref}       
\usepackage{cleveref}
\usepackage{url}            
\usepackage{booktabs}       
\usepackage{amsfonts}       
\usepackage{nicefrac}       
\usepackage{microtype}      
\usepackage{xcolor}         
\usepackage{float}
\usepackage{multirow}
\usetikzlibrary{positioning, arrows.meta, calc, backgrounds, fit}
\usepackage{graphicx}
\usepackage{pifont}  
\newcommand{\cmark}{\textcolor{green!60!black}{\ding{51}}}
\newcommand{\xmark}{\textcolor{red!75!black}{\ding{55}}}

\title{PhysDNet: Physics-Driven Gradient Amplification for Real-Time Image Dehazing }

\author{%
  Rayen Dhahri$^{1}$ \quad
  Lei Sun$^{2}$ \quad
  Victor Javier Kartsch Morinigo$^{3}$ \\
  \textbf{Alice Pagnoux}$^{3}$ \quad
  \textbf{Luca Benini}$^{3, 4}$ \quad
  \textbf{Danda Pani Paudel}$^{2}$ \quad
  \textbf{Luc Van Gool}$^{2}$ \\[0.5em]
  $^{1}$Independent Researcher \\
  $^{2}$INSAIT, Sofia University ``St.~Kliment Ohridski'' \\
  $^{3}$ETH Zurich \\
  $^{4}$Department of Electrical, Electronic and Information Engineering (DEI), University of Bologna
}
\definecolor{cRGB}{HTML}{E8F0FE}
\definecolor{cDep}{HTML}{E6F4EA}
\definecolor{cGate}{HTML}{FEF7E0}
\definecolor{cDec}{HTML}{F3E8FD}
\definecolor{cDir}{HTML}{E8EAED}
\definecolor{cPhy}{HTML}{FCE8E6}

\definecolor{cRGBdark}{HTML}{AECBFA}
\definecolor{cDepdark}{HTML}{A8DAB5}
\definecolor{cGatedark}{HTML}{FDE293}
\definecolor{cDecdark}{HTML}{D7B7FD}
\definecolor{cDirdark}{HTML}{BCC1C6}
\definecolor{cPhydark}{HTML}{F8B4B4}

\definecolor{bgEnc}{HTML}{F8F9FA}
\definecolor{bgDec}{HTML}{FCF7FF}
\definecolor{bgHead}{HTML}{FFFAFA}

\definecolor{rowA}{HTML}{EAF2FB}   
\definecolor{rowB}{HTML}{FDF6EC}   
\definecolor{rowC}{HTML}{EDF7ED}   
\definecolor{rowD}{HTML}{F5EDF8}   
\definecolor{hdrBg}{HTML}{2B579A}  
\definecolor{hdrFg}{HTML}{FFFFFF} 
\definecolor{physBg}{HTML}{FCE8E6} 
\definecolor{dataBg}{HTML}{E8F0FE} 
\definecolor{auxBg}{HTML}{FEF7E0} 
 
\newcommand{\hdr}[1]{\cellcolor{hdrBg}\textcolor{hdrFg}{\textbf{#1}}}

\newcommand{\blfootnote}[1]{%
  \begingroup
  \renewcommand{\thefootnote}{}%
  \footnotetext{#1}%
  \endgroup
}

\begin{document}

\maketitle

\begin{abstract}
Modern dehazing networks are often computationally expensive and overfit to synthetic haze, limiting real-time deployment. This underscores an urgent demand for efficient, high-fidelity restoration methods that can operate on resource-constrained edge devices in practical, dynamic environments. 
We introduce PhysDNet, a lightweight multimodal architecture that fuses NIR imagery with sparse LiDAR depth and is trained with a physics-constrained dual-head objective. Differentiating through a fixed Koschmieder inversion induces \emph{density-adaptive gradient amplification}, acting as an implicit resource allocation mechanism that concentrates learning on heavily degraded regions without increasing inference cost.
PhysDNet delivers superior restoration with minimal capacity and enhanced training efficiency, facilitating real-time deployment on edge devices without increasing inference overhead.
On the Seeing Through Fog benchmark, PhysDNet outperforms competitors by \textbf{1.67 dB} at \textbf{5.2$\times$} lower compute and recovering \textbf{54.9\%} of fog-induced detection loss. Ablations and cross-domain evaluations show that physics-guided training improves generalization including under non-homogeneous haze that violates the Koschmieder assumption. PhysDNet further demonstrates compatibility with fixed-function NPUs, running at \textbf{29/45/99\,FPS} at $<$\textbf{4\,W} across L/M/S variants. Code, deployment scripts and pre-trained checkpoints are available at \url{https://github.com/Heatdh/PhysDnet}.
\end{abstract}
\blfootnote{Correspondence to Rayen Dhahri:  rayendhahri@gmail.com  }

\section{Introduction}



Image dehazing is foundational for outdoor perception under adverse weather, particularly autonomous driving where reduced visibility compromises detection and tracking~\cite{DBLP:conf/cvpr/BijelicGMKRDH20}. Modern autonomous platforms already carry the sensors needed: time-gated NIR cameras that reject nearby backscatter, and LiDAR providing weather-invariant depth~\cite{DBLP:journals/corr/abs-1912-02889}. Active time-gated NIR provides a hardware-level bypass to RGB's failure under dense scattering: synchronized exposure with emitted laser pulses rejects atmospheric backscatter and recovers structural semantics where RGB is effectively blind. Yet most learned dehazing methods~\cite{DBLP:conf/iccv/LiPWXF17,DBLP:conf/aaai/QinWBXJ20,Song_2023,DBLP:journals/tip/ChenHL24, FrDiff2025} process RGB alone and chase quality by scaling capacity, deeper backbones, wider attention, iterative refinement, inflating inference cost and relying on operators such as self-attention and layer normalization, which are incompatible with fixed-function edge accelerators~\cite{dhahri2025quanttrimpracticeimprovedcrossplatform}. 
Consequently, a practical dehazing solution must be lightweight enough for deployment on low-power edge accelerators while seamlessly exploiting the multi-modal sensor suites already ubiquitous on autonomous platforms. 

We introduce \textbf{PhysDNet}, a lightweight multimodal dehazing
network for real-time edge deployment. PhysDNet fuses gated NIR
imagery with sparse LiDAR depth through mask-aware multi-scale gating
and is trained with a \emph{physics-constrained dual-head framework}:
a deployable direct restoration head and a parameter-free auxiliary
head that inverts the Koschmieder scattering
model~\cite{mccartney1976optics}.

Our central insight is that the auxiliary physics head improves
learning not by constraining the final output, but by shaping the
\emph{optimization dynamics}. Backpropagating through the fixed
physical inversion induces transmission-dependent gradient scaling
proportional to $(t+\epsilon)^{-2}$, automatically amplifying
supervision in heavily degraded regions. We call this
\emph{density-adaptive gradient amplification}: a hyperparameter-free
implicit curriculum that prioritizes the hardest pixels during
training at zero inference cost. \textbf{To our knowledge, this is the first analysis of differentiation through a physical inversion as a mechanism for shaping optimization via structured gradient scaling.}
This mechanism improves performance through more efficient use of gradients during training rather than increased model capacity. In contrast to scaling architectures, which increases inference cost, PhysDNet achieves its gains by reallocating optimization focus, leaving the deployment footprint unchanged.




We deliver the following contributions:

\begin{itemize}
    \item \textbf{Density-adaptive gradient amplification.} A dual-head training framework where a fixed inversion induces $(t+\epsilon)^{-2}$ gradient scaling, focusing learning on degraded regions without explicit weighting or extra inference cost.

    \item \textbf{State-of-the-art accuracy at low compute.} On STF, PhysDNet reaches 36.46\,dB PSNR at 18.98\,GMACs, outperforming FFA-Net by $+1.67$\,dB at $5.2\times$ lower compute, and recovers 54.9\% of YOLOv8 detection loss under aggressive fog.

    \item \textbf{Edge deployability.} The only evaluated method combining competitive restoration quality with fixed-function NPU compatibility, running at 29-99\,FPS at $<$4\,W under standard INT8 quantization.
\end{itemize}

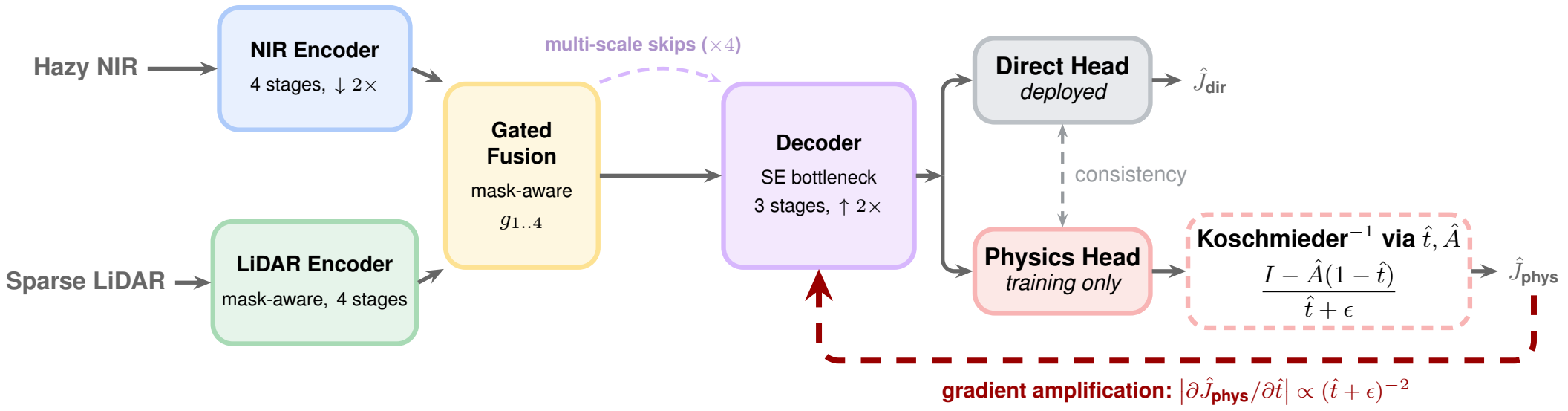
\begin{figure}[t]
\centering

\begin{tikzpicture}[
    >=Stealth,
    font=\sffamily\small,
    transform shape, scale=0.68,
    node distance=0.6cm,
    encBlock/.style={draw=cRGBdark, line width=1.2pt, fill=cRGB,
                     rounded corners=5pt, minimum width=2.5cm,
                     minimum height=1.6cm, align=center,
                     font=\sffamily\footnotesize\bfseries},
    encBlockD/.style={draw=cDepdark, line width=1.2pt, fill=cDep,
                      rounded corners=5pt, minimum width=2.5cm,
                      minimum height=1.6cm, align=center,
                      font=\sffamily\footnotesize\bfseries},
    fuseBlock/.style={draw=cGatedark, line width=1.2pt, fill=cGate,
                      rounded corners=5pt, minimum width=2.0cm,
                      minimum height=2.4cm, align=center,
                      font=\sffamily\footnotesize\bfseries},
    decBlock/.style={draw=cDecdark, line width=1.2pt, fill=cDec,
                     rounded corners=5pt, minimum width=2.5cm,
                     minimum height=2.4cm, align=center,
                     font=\sffamily\footnotesize\bfseries},
    headDir/.style={draw=cDirdark, line width=1.2pt, fill=cDir,
                    rounded corners=5pt, minimum width=2.3cm,
                    minimum height=1.1cm, align=center,
                    font=\sffamily\small\bfseries},
    headPhy/.style={draw=cPhydark, line width=1.2pt, fill=cPhy,
                    rounded corners=5pt, minimum width=2.3cm,
                    minimum height=1.1cm, align=center,
                    font=\sffamily\small\bfseries},
    pill/.style={draw=black!40, line width=0.8pt, fill=white,
                 rounded corners=5pt, inner sep=3pt,
                 font=\sffamily\footnotesize\bfseries},
    flow/.style={-{Stealth[length=4.5pt,width=3.5pt]},
                 line width=1pt, color=black!60, rounded corners=3pt},
    skipflow/.style={-{Stealth[length=3.5pt,width=2.5pt]},
                     line width=0.9pt, color=cDecdark, densely dashed,
                     rounded corners=4pt},
    consflow/.style={{Stealth[length=3.5pt]}-{Stealth[length=3.5pt]},
                     line width=0.8pt, color=cDirdark!80!black, densely dashed},
    gradflow/.style={-{Stealth[length=8pt,width=7pt]},
                     line width=1.5pt, color=red!60!black, dashed,
                     dash pattern=on 5pt off 4pt, rounded corners=5pt},
    seclabel/.style={font=\sffamily\small\bfseries, text=black!60}
]

\node[seclabel] (inNIR) at (0, 1.4) {Hazy NIR};
\node[seclabel] (inLID) at (0, -1.4) {Sparse LiDAR };

\node[encBlock] (encR) at (3.0, 1.4)
   {NIR Encoder\\[3pt]
    \scriptsize\mdseries 4 stages,\, $\downarrow 2\times$};
\node[encBlockD] (encD) at (3.0, -1.4)
   {LiDAR Encoder\\[3pt]
    \scriptsize\mdseries mask-aware,\, 4 stages};

\draw[flow] (inNIR) -- (encR);
\draw[flow] (inLID) -- (encD);

\node[fuseBlock] at ($(encR.east)!0.5!(encD.east) + (1.4, 0)$) (fuse)
   {Gated\\Fusion\\[3pt]
    \scriptsize\mdseries mask-aware\\[1pt]
    $g_{1..4}$};
\draw[flow] (encR.east) -- (fuse.north west);
\draw[flow] (encD.east) -- (fuse.south west);

\node[decBlock, right=1.6cm of fuse] (dec)
   {Decoder\\[3pt]
    \scriptsize\mdseries SE bottleneck\\[1pt]
    \scriptsize\mdseries 3 stages,\, $\uparrow 2\times$};

\draw[flow] (fuse) -- (dec);
\draw[skipflow] (fuse.north east) to[bend left=22]
  node[above=2pt, font=\sffamily\scriptsize\bfseries, text=cDecdark!80!black, pos=0.35]
  {multi-scale skips ($\times 4$)}
  (dec.north west);

\coordinate (split) at ($(dec.east) + (0.35, 0)$);
\draw[flow] (dec.east) -- (split);

\node[headDir, right=0.4cm of split, yshift=1.25cm] (hDir)
   {Direct Head\\[-1pt]
    \footnotesize\mdseries\itshape deployed};
\draw[flow] (split) |- (hDir.west);

\node[headPhy, right=0.4cm of split, yshift=-1.25cm] (hPhy)
   {Physics Head\\[-1pt]
    \footnotesize\mdseries\itshape training only};
\draw[flow] (split) |- (hPhy.west);

\node[seclabel, right=0.4cm of hDir] (outDir) {$\hat{J}_{\text{dir}}$};
\draw[flow] (hDir) -- (outDir);

\node[headPhy, draw=cPhydark, densely dashed, fill=white,
      right=0.45cm of hPhy,
      minimum width=3.0cm, minimum height=1.4cm] (inv)
   {Koschmieder$^{-1}$ via $\hat{t}, \hat{A}$\\[2pt]
    \small\mdseries $\dfrac{I - \hat{A}(1-\hat{t})}{\hat{t}+\epsilon}$};
\draw[flow] (hPhy.east) -- (inv.west);

\node[seclabel, right=0.35cm of inv] (outPhy) {$\hat{J}_{\text{phys}}$};
\draw[flow] (inv.east) -- (outPhy.west);

\draw[consflow] (hDir.south) -- 
  node[midway, right=1pt, font=\sffamily\footnotesize, text=cDirdark!80!black] {consistency} 
  (hPhy.north);

\path ($(dec.south) + (0, -1.2)$) coordinate (gradTarget);
\draw[gradflow] (outPhy.south) -- (outPhy.south |- gradTarget) -- 
  node[below=2pt, midway, font=\sffamily\footnotesize\bfseries, text=red!60!black, align=center] 
    {gradient amplification:\ $\bigl|\partial \hat{J}_{\text{phys}}/\partial \hat{t}\bigr| \propto (\hat{t}+\epsilon)^{-2}$}
  (gradTarget) -- (dec.south);
\end{tikzpicture}

\caption{\textbf{PhysDNet architecture.} Dual encoders extract NIR and sparse-depth features, merged by mask-aware gating with Squeeze-and-Excitation at the bottleneck. The shared representation feeds a direct head (deployed at inference) and a physics head (training only). Backpropagating through the Koschmieder inversion scales gradients by $(\hat{t}+\epsilon)^{-2}$, concentrating optimization on heavily degraded regions at zero inference cost.}
\label{fig:architecture}
\vspace{-1em}
\end{figure}

\section{Method}
We adopt the Koschmieder scattering model~\cite{mccartney1976optics}: for a pixel $x$, the hazy intensity $I(x)$ is a convex combination of scene radiance $J(x)$ and global airlight $A$,
\begin{equation}
  I(x) = J(x)\,t(x) + A\bigl(1 - t(x)\bigr),
  \qquad t(x) = e^{-\beta d(x)},
  \label{eq:koschmieder}
\end{equation}
with transmission $t(x)\in(0,1]$, scattering coefficient $\beta$, and depth $d(x)$. Single-image recovery is ill-posed in both $t$ and $A$~\cite{he2010single}; PhysDNet breaks this ambiguity by fusing sparse LiDAR depth (weather-invariant) with time-gated NIR imagery (rejects nearby backscatter)~\cite{DBLP:journals/corr/abs-1912-02889}. Extended background in Appendix~\ref{app:ext_background}.

We present \textbf{PhysDNet} (\textbf{Phys}ics \textbf{D}ual-head
Dehazing \textbf{Net}work), a lightweight architecture for
LiDAR-guided image dehazing. A fixed, non-learnable physics inversion
head operates in parallel with a data-driven restoration head, both
sharing a common encoder-decoder backbone. Backpropagation through the physics inversion induces $(\hat{t}+\epsilon)^{-2}$ gradient scaling, concentrating optimization on severely degraded regions without additional parameters.

The model receives as input a hazy image $I \in [0,1]^{3 \times H \times W}$,
sparse LiDAR depth $D_s \in \mathbb{R}^{1 \times H \times W}$,
and a binary validity mask $M \in \{0,1\}^{1 \times H \times W}$,
and jointly predicts a restored image, a per-pixel transmission map,
and a global airlight vector.

\subsection{Network Architecture}\label{sec:architecture}

PhysDNet is a dual-branch encoder-decoder (\cref{fig:architecture}) with gated cross-modal fusion and two output heads. Two modality-specific encoder branches process NIR and depth independently at four spatial scales, merging features at each scale via a learned gate. A single decoder upsamples the fused representations. Unlike a standard U-Net~\cite{DBLP:conf/miccai/RonnebergerFB15}, each branch learns modality-specific representations before fusion, with gating adapting to local signal availability.

\textbf{Dual encoder and decoder.}
Two modality-specific encoder branches follow a four-stage downsampling structure using depthwise-separable (DWSep) convolutions~\cite{howard2017mobilenetsefficientconvolutionalneural}, BatchNorm, and ReLU6~\cite{Sheng_2018}, with residual shortcuts and $2{\times}$ max-pooling per stage for edge efficiency. The image branch accepts 3-channel NIR input; the depth branch receives the sparse depth $D_s$ concatenated with the validity mask $M$. A Squeeze-and-Excitation (SE) channel attention block~\cite{DBLP:journals/corr/abs-1709-01507} is applied once at the bottleneck after gated fusion. The decoder applies three bilinear upsampling stages with skip connections from each encoder scale. Full layer-wise specifications are in Appendix~\ref{app:arch_details}.

\textbf{Gated fusion.}
At each of the four encoder scales, the image features
$\mathbf{f}_{\text{img}}$ and depth features
$\mathbf{f}_{\text{depth}}$ are combined via a learned gate:
\begin{equation}
  g = \sigma\!\bigl(W_g [\mathbf{f}_{\text{img}};\,\mathbf{f}_{\text{depth}}]\bigr),
  \qquad
  \mathbf{f}_{\text{fused}} = g \odot \mathbf{f}_{\text{img}}
    + (1 - g) \odot \mathbf{f}_{\text{depth}},
  \label{eq:gated_fusion}
\end{equation}
where $[\cdot\,;\cdot]$ denotes channel-wise concatenation and
$W_g$ is a $1{\times}1$ convolution  following the gated multimodal fusion paradigm \cite{10.1007/s00521-019-04559-1, DBLP:conf/cvpr/ChengCLZH17}.  This mechanism is inherently
mask-aware: where LiDAR returns are absent ($M{=}0$), the depth
features carry no geometric signal and the gate learns to rely on
the image branch; where depth is present, geometric cues help
disambiguate haze density from scene content.

\textbf{Model family.}
Scaling the base channel width $c \in \{32, 64, 96\}$ yields three variants (S/M/L) spanning the accuracy, efficiency Pareto front (\cref{tab:variants}). Efficiency metrics are presented in \cref{tab:main_comparison}; even PhysDNet-L (18.98 GMACs) remains lower-cost than evaluated baselines.
 
\subsection{Physics-Constrained Dual-Head Design}
\label{sec:heads}
 
The shared decoder features
$\mathbf{z} \in \mathbb{R}^{c \times H \times W}$ feed two
complementary output heads.  This dual-head design is the core
mechanism through which physical knowledge is injected into the
learning process.
 
\textbf{Direct Head (data-driven).}
A convolutional projection with sigmoid activation maps decoder features directly to the image domain, yielding $\hat{J}_{\mathrm{dir}}(x)\in[0,1]^3$. This unconstrained estimator makes no assumption about the physical image formation process and serves as the deployed output, providing robustness against Koschmieder-model violations (spatially varying airlight, multiple scattering, sensor artifacts).
 
\textbf{Physics Head (structured).}
Two auxiliary sub-heads predict the atmospheric scattering parameters
from the same shared features:
\begin{align}
  \hat{t}(x) &= t_{\min} + (t_{\max} - t_{\min})\,
    \sigma\!\bigl(W_t * \mathbf{z}(x) + b_t\bigr)
    \;\in\; [t_{\min},\,t_{\max}],
  \label{eq:head_trans} \\[2pt]
  \hat{A} &= \sigma\!\bigl(W_A \cdot \operatorname{GAP}(\mathbf{z}_{\text{bot}})
    + b_A\bigr)
    \;\in\; [0,1]^3,
  \label{eq:head_airlight}
\end{align}
where $W_t$ is a $1{\times}1$ convolution producing per-pixel
transmission, $W_A$ is a fully connected layer applied after global
average pooling of the bottleneck features $\mathbf{z}_{\text{bot}}$,
and $t_{\min}{=}0.05$, $t_{\max}{=}0.95$ prevent numerical
instability.
 
The physics-restored image is obtained by inverting
Eq.~\eqref{eq:koschmieder} via its closed-form solution:
\begin{equation}
  \hat{J}_{\text{phys}}(x) = \operatorname{clamp}\!\left(
    \frac{I(x) - \hat{A}\,(1 - \hat{t}(x))}{\hat{t}(x) + \epsilon},\;
    0,\;1
  \right),
  \label{eq:physics_inversion}
\end{equation}
with $\epsilon = 10^{-4}$.  This inversion contains \textbf{zero
learnable parameters}: it is a fixed, differentiable function whose
output quality is determined solely by the accuracy of $\hat{t}$ and
$\hat{A}$.  During training, gradients flow backwards through this
fixed computation into the shared encoder, producing the structured
gradient signal analysed in~\S\ref{sec:gradient}.

\textbf{Inference behaviour.}
At test time only the direct head is deployed; the physics head operates solely during training. This ensures reported gains reflect improved features rather than test-time physics exploitation.

\subsection{Density-Adaptive Gradient Amplification}
\label{sec:gradient}
 
We now derive the gradient signal the physics head propagates into the shared encoder. The structure reveals an implicit curriculum that emerges from the scattering inversion itself, without explicit
sample weighting or scheduling.

Taking partial derivatives of Eq.~\eqref{eq:physics_inversion}
(within the valid range, before clamping) with respect to the
predicted scattering parameters:

\noindent
\begin{minipage}{0.5\linewidth}
\begin{equation}
  \frac{\partial \hat{J}_{\text{phys}}}{\partial \hat{t}} = \frac{\hat{A} - I(x)}{(\hat{t} + \epsilon)^2},
  \label{eq:grad_t}
\end{equation}
\end{minipage}%
\begin{minipage}{0.5\linewidth}
\begin{equation}
  \frac{\partial \hat{J}_{\text{phys}}}{\partial \hat{A}} = \frac{\hat{t} - 1}{\hat{t} + \epsilon},
  \label{eq:grad_A}
\end{equation}
\end{minipage}
\vspace{1ex}

\noindent where we neglect the $O(\epsilon)$ term in the numerator of
Eq.~\eqref{eq:grad_t}. In practice, $\epsilon=10^{-4}$ remains several orders of magnitude smaller than $t_{\min}=0.05$, ensuring the approximation holds even in the lowest-transmission regime.

\textbf{Transmission gradient: implicit hard-example mining.}
The factor $(\hat{t} + \epsilon)^{-2}$ in Eq.~\eqref{eq:grad_t}
implies that gradient magnitude scales \emph{inversely with the square of transmission}. In dense fog ($\hat{t} \to t_{\min} = 0.05$), gradients are amplified by up to $(t_{\min} + \epsilon)^{-2} \approx 400\times$ relative to clear-air pixels ($\hat{t} \approx 1$), concentrating learning on heavily degraded regions. This scaling is empirically confirmed in \cref{fig:grad_amp}(a): PhysDNet-M follows the $(\hat{t}+\epsilon)^{-2}$ curve and exceeds the non-physics baseline by up to ${\sim}694\times$ in dense haze (\cref{fig:grad_amp}(b)).

This behaviour acts as an implicit resource allocation mechanism during training, analogous to hard-example mining~\cite{shrivastava2016trainingregionbasedobjectdetectors}, where degraded pixels dominate the gradient signal. Unlike explicit weighting schemes (e.g., focal loss~\cite{lin2018focallossdenseobject}), this effect requires \emph{no additional hyperparameters} and arises directly from differentiating through the physics inversion.

\textbf{Variance correspondence.}
The same $t^{-2}$ structure governs the inversion's noise variance, $\operatorname{Var}[\hat{J}_{\mathrm{phys}}] = \sigma^2/t^2 + O(\delta)$ (Appendix~\ref{sec:theory_proof}, Prop.~1), making the physics head the structural inverse of classical heteroscedastic regression~\cite{mai2021batchinversevarianceweightingdeep}, which \emph{down}-weights high-variance samples. Full derivations, airlight-gradient analysis, bounded-gradient discussion, and per-pixel spatial validation on real STF fog are in Appendix~\ref{sec:theory_proof}.
Empirically, on 300 STF validation frames with known synthetic sensor noise ($\sigma{\approx}0.01$), the squared error of the inversion binned by ground-truth transmission follows $\sigma^2/t^2$ across two orders of magnitude (log-log $r{=}0.981$, 24 bins, clamp-saturated pixels excluded; App.~\ref{app:prop1_validation}).
\vspace{-0.5em}

\textbf{Scope of the mechanism.}
The inversion induces transmission-dependent Jacobian scaling, but its restoration benefit depends on whether the inverted parameter is \emph{anchored} during training. Without transmission supervision (RGB-only RESIDE-6K~\cite{li2019benchmarking}, LOL-v1~\cite{wei2018retinex}), physics and NoPhy variants are within 0.15\,dB. Where a measurement anchors the parameter, the gain reappears beyond dehazing: depth-anchored underwater restoration ($+0.38$\,dB overall, $+0.82$\,dB mid-transmission) and Retinex~\cite{land1977retinex} low-light enhancement on LOL-v2-Real~\cite{yang2021lolv2} ($+0.5$--$0.6$\,dB), the $A{\to}0$ case of Eq.~\eqref{eq:physics_inversion} (App.~\ref{app:rgb_scope}). The anchored STF setting yields the largest gain (\S\ref{sec:physics_ablation}).

\subsection{Physics-Anchored Residual Learning}
\label{sec:loss}

The physics-reconstruction loss $\mathcal{L}_{\text{pr}}$ anchors the
shared latent representation in a physically meaningful space, while
the standard reconstruction loss $\mathcal{L}_{\text{rec}}$ lets the
direct head learn residual corrections for non-homogeneous fog and
sensor effects beyond Koschmieder, ensuring the network is
physics-anchored but not physics-limited. A contrastive term $\mathcal{L}_{\text{contra}}$ pushes Direct Head features away from the hazy input, activated late in training. The full objective is organized in three families (data-driven, physics, auxiliary) with every equation in App.~\ref{app:losses}.

\section{Experiments}
\label{sec:experiments}
    
We evaluate PhysDNet along three axes. First, \textbf{quality} on STF 
(\S\ref{sec:base}--\S\ref{sec:physics_ablation}), out-of-distribution fog 
(\S\ref{sec:beta_sweep}), and real STF fog (\S\ref{sec:real_world}). Second, 
\textbf{utility} for downstream detection (\S\ref{sec:downstream_yolo}) and 
zero-shot cross-domain transfer (\S\ref{sec:ohaze}). Third, \textbf{deployability} 
on two automotive edge platforms (\S\ref{sec:edge_deployment}). Appendices 
report LiDAR-sparsity robustness (App.~\ref{app:lidar_sparsity}) and 
no-reference perceptual quality (App.~\ref{app:niqe}).
 
\subsection{Experimental Setup}
\label{sec:setup}

We train and evaluate on the Seeing Through Fog (STF) dataset~\cite{DBLP:conf/cvpr/BijelicGMKRDH20}, using 4\,212 training frames and 1\,299 validation frames with on-the-fly synthetic haze generated via Eq.~\eqref{eq:koschmieder}. Sparse LiDAR is used unmodified at both train and test time. Dataset details in App.~\ref{app:stf_details}.

We compare against Dark Channel Prior (DCP)~\cite{DBLP:conf/cvpr/HeST09}, AOD-Net~\cite{DBLP:conf/iccv/LiPWXF17}, FFA-Net~\cite{DBLP:conf/aaai/QinWBXJ20}, DehazeFormer-T~\cite{Song_2023}, and DEA-Net~\cite{DBLP:journals/tip/ChenHL24}. All PhysDNet variants are trained with AdamW at $256{\times}256$ using only NPU/TensorRT-friendly operators (depthwise-separable convolutions, BatchNorm, ReLU6, bilinear upsampling). All reported results correspond to the unconstrained direct head, ensuring gains cannot stem from trivial consistency with the synthetic generation process. Recent heavyweight (UVM-Net~\cite{uvm}, DFFNet~\cite{DFFNet}, Ksformer~\cite{ksformer}) and diffusion-based (HazeDiff~\cite{lin2025hazediff}, DehazeDDPM~\cite{yu2023dehazeddpm}, FrDiff~\cite{FrDiff2025}) methods are excluded as they exceed real-time edge constraints; selection rationale in App.~\ref{app:model_selection}.

\subsection{Restoration Quality on STF}
\label{sec:base}

\begin{figure}
  \centering
  \includegraphics[width=\linewidth]{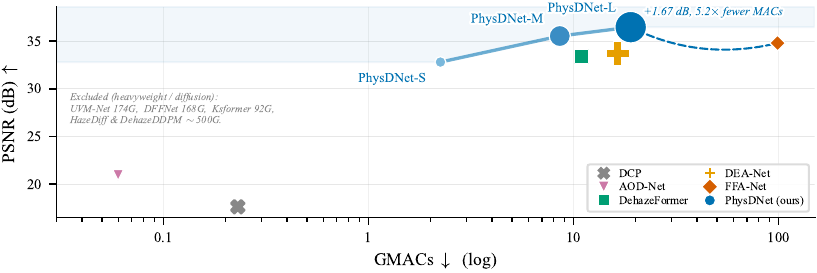}
  \vspace{-2em} 
\caption{%
    \textbf{Accuracy vs.\ computational cost on STF.} PhysDNet establishes a new Pareto front in the edge-deployable region. Baselines are plotted at their best-performing resolution (Table~\ref{tab:main_comparison}). PhysDNet-L outperforms FFA-Net by $+1.67$\,dB at $5.2\times$ lower compute; PhysDNet-S retains competitive accuracy at ultra-low compute. Aside from PhysDNet, AOD-Net compiles to the NPU but its restoration quality is insufficient for deployment. Heavyweight and diffusion baselines excluded (App.~\ref{app:model_selection}).}
  \label{fig:gmacs_vs_psnr}
  \vspace{-0.5em} 
\end{figure}

\Cref{tab:main_comparison} compares PhysDNet against four learned baselines and the classical DCP~\cite{DBLP:conf/cvpr/HeST09}. At matched $256{\times}256$ resolution, PhysDNet-L reaches 36.46\,dB PSNR, outperforming FFA-Net+LiDAR at its preferred $512{\times}512$ resolution (34.79\,dB) by $+1.67$\,dB at $5.2\times$ lower compute. PhysDNet-M (2.95\,M, 8.56\,GMACs) surpasses DEA-Net by $+1.82$\,dB at roughly half the cost. The gap persists across resolutions, and \S\ref{sec:physics_ablation} confirms the physics-constrained training regime, not backbone capacity, is the primary driver. Qualitative comparisons in Appendix~\ref{app:qualitative}.

\begin{table}
\centering
\caption{%
  \textbf{Quantitative comparison on the STF validation set.}
  PSNR/SSIM for the \emph{direct} restoration head. PhysDNet-M reports mean$\pm$SE over 3 seeds.
  Best in \textbf{bold}, second \underline{underlined}. GMACs at the indicated resolution.
}
\label{tab:main_comparison}
\small
\setlength{\tabcolsep}{4pt}
\renewcommand{\arraystretch}{0.95}
\begin{tabular}{llcccccc}
\toprule
\textbf{Method} & \textbf{Input} & \textbf{Res.}
  & \textbf{Params\,(M)} & \textbf{GMACs}
  & \textbf{PSNR\,(dB)\,$\uparrow$} & \textbf{SSIM\,$\uparrow$} \\
\midrule
DCP~\cite{DBLP:conf/cvpr/HeST09}          & NIR           & 256 & 0     &   ---  & 17.62 & 0.5992 \\
\midrule
AOD-Net~\cite{DBLP:conf/iccv/LiPWXF17}     & NIR           & 512 & 0.002 &   0.23 & 18.62 & 0.7123 \\
AOD-Net                                    & NIR           & 256 & 0.002 &   0.06 & 20.69 & 0.7499 \\
AOD-Net                                    & LiDAR+Gated   & 512 & 0.002 &   0.24 & 17.05 & 0.6669 \\
AOD-Net                                    & LiDAR+Gated   & 256 & 0.002 &   0.06 & 20.95 & 0.7329 \\
\midrule
DehazeFormer-T~\cite{Song_2023}            & NIR           & 512 & 1.36  &  10.92 & 33.38 & 0.9348 \\
DehazeFormer-T                             & NIR           & 256 & 1.36  &   2.73 & 32.22 & 0.9398 \\
DehazeFormer-T                             & NIR+LiDAR     & 512 & 1.36  &  10.97 & 29.91 & 0.9092 \\
\midrule
FFA-Net~\cite{DBLP:conf/aaai/QinWBXJ20}    & NIR           & 512 & 0.77  &  98.57 & 33.90 & 0.9474 \\
FFA-Net                                    & NIR           & 256 & 0.77  &  24.64 & 33.80 & 0.9504 \\
FFA-Net                                    & NIR+LiDAR     & 512 & 0.77  &  98.73 & 34.79 & 0.9516 \\
FFA-Net                                    & NIR+LiDAR     & 256 & 0.77  &  24.68 & 33.23 & 0.9476 \\
\midrule
DEA-Net~\cite{DBLP:journals/tip/ChenHL24}  & NIR           & 256 & 3.65  &  16.42 & 33.71 & \underline{0.9588} \\
DEA-Net                                    & LiDAR         & 256 & 3.65  &  16.44 & 33.53 & 0.9572 \\
\midrule
PhysDNet-S                                 & NIR+LiDAR     & 256 & 0.72  &   2.24 & 32.80 & 0.9331 \\
PhysDNet-M                                 & NIR+LiDAR     & 256 & 2.95  &   8.56
  & \underline{35.55{\scriptsize$\pm$0.07}}  & 0.9566{\scriptsize$\pm$0.0009} \\
PhysDNet-L                                 & NIR+LiDAR     & 256 & 6.59  &  18.98
  & \textbf{36.46} & \textbf{0.9606} \\
\bottomrule
\end{tabular}
\end{table}
\subsection{Physics Head Ablation}
\label{sec:physics_ablation}

To isolate the effect of physics-constrained training, we remove the
physics head (\emph{NoPhy}) from an identical ch64 model, disabling
transmission/airlight prediction and all physics losses. This reduces
parameters by only 0.04\,M (2.95$\to$2.91\,M) but causes a large drop
in performance (Table~\ref{tab:physics_ablation}).

\begin{table}[H]
\centering
\caption{\textbf{Physics head ablation.} Removing the physics head degrades PSNR by 5.51\,dB and reduces downstream mAP recovery.}
\label{tab:physics_ablation}
\resizebox{\linewidth}{!}{
\begin{tabular}{lcccccc}
\toprule
\textbf{Variant} & \textbf{Physics Head}
  & \textbf{Params\,(M)}
  & \textbf{PSNR\,(dB)}
  & \textbf{SSIM}
  & \textbf{mAP\,(\%)}
  & \textbf{$\Delta$\,mAP} \\
\midrule
PhysDNet-M        & \checkmark & 2.95 & \textbf{35.55} & \textbf{0.9569} & \textbf{11.70} & \textbf{+3.05} \\
PhysDNet-M NoPhy  & \texttimes & 2.91 & 30.02 & 0.9263 & 10.70 & +2.06 \\
\midrule
\multicolumn{2}{l}{\emph{Difference}} & $-$0.04 & $-$5.51 & $-$0.0306 & $-$1.00 & $-$0.99 \\
\bottomrule
\end{tabular}}
\vspace{-8pt}
\end{table}

Removing the physics head reduces PSNR by \textbf{5.51\,dB} and mAP
recovery from $+3.05$ to $+2.06$, indicating that gains arise from the training
regime rather than architecture alone.

\begin{figure}[t]
  \centering
  \includegraphics[width=\linewidth]{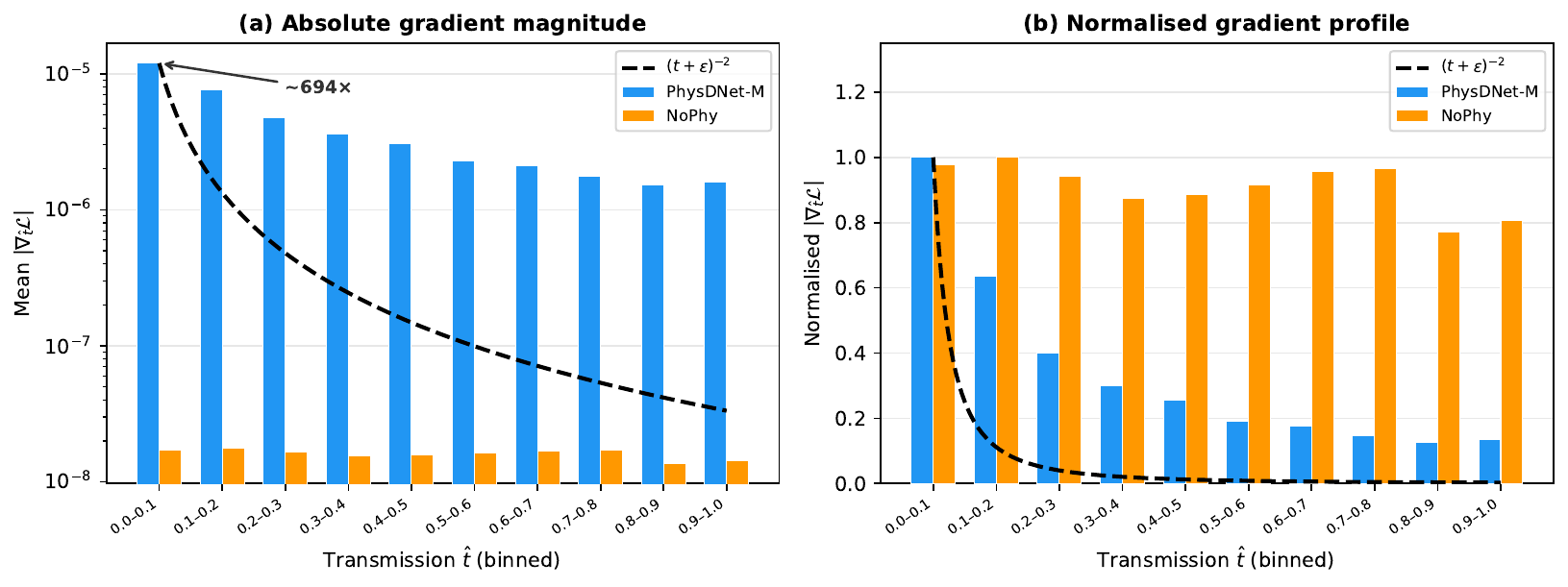}
  \vspace{-2em} 
  \caption{
    \textbf{Density-adaptive gradient amplification.}
    PhysDNet-M follows the $(t+\epsilon)^{-2}$ scaling and concentrates
    gradients on low-transmission regions, unlike the flat NoPhy baseline.}
  \label{fig:grad_amp}
  \vspace{-1em}
\end{figure}

\textbf{Gradient amplification.}
\Cref{fig:grad_amp} confirms the $(\hat{t}+\epsilon)^{-2}$ scaling:
PhysDNet-M gradients follow the theoretical curve and exceed NoPhy
by up to ${\sim}694\times$ in dense haze, reflecting implicit
hard-example mining.

\subsection{Out-of-Distribution Fog Robustness}
\label{sec:beta_sweep}

We sweep $\beta\in[0.005, 0.10]$ at $A{=}0.8$ on the full validation set (1\,299 frames per $\beta$), with no retraining (see Appendix~\ref{app:beta_sweep} for the full performance curve); values $\beta \geq 0.05$ lie outside the training range.
PhysDNet-M maintains the highest PSNR across the entire range and
degrades smoothly beyond the training boundary, retaining a
$>$6\,dB advantage over FFA-Net at $\beta{=}0.06$.
We attribute this robustness to the physics-constrained encoder:
because the inversion formula (Eq.~\ref{eq:physics_inversion}) is
domain-invariant, the learned $(t, A, J)$ factorisation generalises
to unseen $\beta$ values without retraining.

\subsection{Downstream Object Detection}
\label{sec:downstream_yolo}

Pixel-level metrics on synthetic data do not guarantee that
semantically meaningful structure has been recovered.
We therefore measure the impact of dehazing on downstream object
detection using YOLOv8-M~\cite{yaseen2024yolov8indepthexplorationinternal}
(pretrained on COCO, conf$\geq$0.25, IoU$\geq$0.5) on STF
validation frames with synthetically added \emph{aggressive} fog
($\beta{=}0.06$, out-of-distribution).
Full protocol details are in \cref{app:detection_protocol}. We report relative recovery rather than absolute performance, as the detector is not fine-tuned for STF.
 
\begin{table}[t]
\centering
\caption{%
  \textbf{Downstream object detection} on 426 STF val frames with synthetic fog ($\beta{=}0.06$). Detector: YOLOv8-M (conf$\geq$0.25, IoU$\geq$0.5). $\Delta$mAP is relative to the foggy baseline; recovery = $\Delta\text{mAP}/|\text{fog degradation}|$.
}
\label{tab:downstream}
\small
\setlength{\tabcolsep}{4pt}
\renewcommand{\arraystretch}{0.92}
\begin{tabular}{lccccc}
\toprule
\textbf{Condition} & \textbf{AP\textsubscript{veh}} & \textbf{AP\textsubscript{pers}}
  & \textbf{AP\textsubscript{cyc}} & \textbf{mAP (\%)} & \textbf{$\Delta$mAP} \\
\midrule
Clean (upper bound)        & 33.95 & 7.97 & 1.05 & 14.33 & ---   \\
Foggy ($\beta{=}0.06$)     & 18.09 & 5.42 & 2.42 &  8.64 & $-$5.68 \\
\midrule
AOD-Net (NIR)              & 15.14 & 4.66 & 1.27 &  7.02 & $-$1.62 \\
AOD-Net (LiDAR+Gated)      & 14.40 & 4.23 & 1.49 &  6.70 & $-$1.94 \\
DehazeFormer-T (NIR)       & 16.37 & 5.74 & 1.65 &  7.92 & $-$0.72 \\
DehazeFormer-T (LiDAR)     & 15.62 & 5.24 & 1.21 &  7.36 & $-$1.29 \\
FFA-Net (NIR)              & 24.37 & 6.30 & 1.32 & 10.66 & $+$2.02 \\
FFA-Net (LiDAR)            & 23.09 & 5.35 & 1.39 &  9.94 & $+$1.30 \\
DEA-Net (NIR)              & 25.82 & 5.82 & 1.66 & 11.10 & $+$2.45 \\
DEA-Net (LiDAR)            & 25.52 & 5.72 & 2.00 & 11.08 & $+$2.44 \\
\midrule
PhysDNet-S                 & 24.49 & 5.94 & 1.22 & 10.55 & $+$1.91 \\
\textbf{PhysDNet-M}        & \textbf{27.65} & 6.45 & 0.99 & \textbf{11.70} & $+$\textbf{3.05} \\
\textbf{PhysDNet-L}        & 27.60 & \textbf{6.70} & 0.99 & \textbf{11.76} & $+$\textbf{3.12} \\
\midrule
\multicolumn{6}{@{}l}{\footnotesize Recovery: PhysDNet-L=\textbf{54.9\%}, PhysDNet-M=53.7\%, DEA-Net (NIR)=43.1\%, FFA-Net (NIR)=35.6\%}\\
\bottomrule
\end{tabular}
\vspace{-0.5em}
\end{table}
 
\begin{figure}[H]
  \centering
  {\small
  \begin{tabular}{@{}p{0.245\linewidth}@{\hspace{2pt}}p{0.245\linewidth}@{\hspace{2pt}}p{0.245\linewidth}@{\hspace{2pt}}p{0.245\linewidth}@{}}
    \centering Foggy Image &
    \centering FFA-Net &
    \centering PhysDNet-M &
    \centering\arraybackslash Clean Image
  \end{tabular}}\\[-2pt]
  \includegraphics[width=\linewidth]{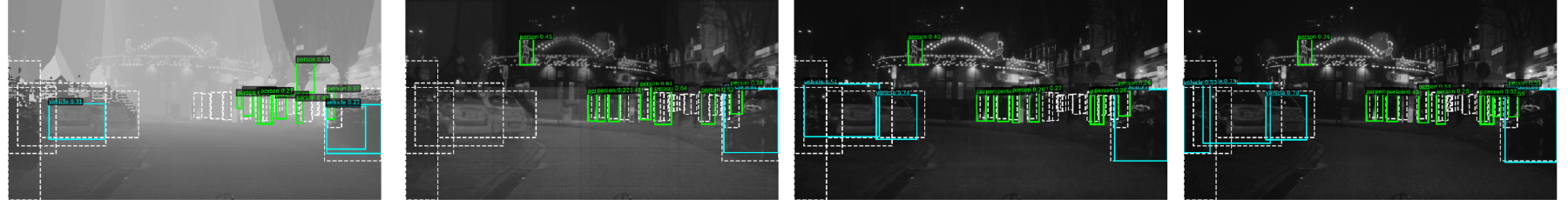}
  \vspace{-1em}
  \caption{%
    \textbf{Downstream detection after dehazing}
    ($\beta{=}0.06$).
    PhysDNet-M recovers scene contrast and enables more
    accurate detections compared to FFA-Net.
    Dashed white: ground truth; solid coloured: predictions.
  }
  \vspace{-1.5em}
  \label{fig:qualitative}
  
\end{figure}

PhysDNet-L recovers 3.12\,pp (\textbf{54.9\%} of the degradation) and PhysDNet-M 3.05\,pp (53.7\%), while the best modern baseline, DEA-Net (NIR), recovers only 2.45\,pp (43.1\%). This indicates that PhysDNet restores semantically relevant structure rather than only improving pixel-level metrics. A representative qualitative example is shown in \cref{fig:qualitative}.
 
\subsection{Zero-Shot Cross-Domain Generalization and Non-Homogeneous Fog}
\label{sec:ohaze}
We evaluate zero-shot on \textbf{O-HAZE}~\cite{ancuti2018ohazedehazingbenchmarkreal} and \textbf{I-HAZE}~\cite{DBLP:conf/acivs/AncutiATV18}, imposing a quadruple shift (NIR$\to$RGB, synthetic$\to$real, LiDAR$\to$none, outdoor$\to$indoor). No fine-tuning. Due to near-achromatic outputs of the monochrome-trained model, we report luminance (Y) metrics following prior work; colour is learned via adaptation in \S\ref{sec:rgb_adaptation}.

\begin{table}
\centering
\caption{\textbf{Zero-shot cross-domain generalization} on O-HAZE and I-HAZE. All models trained on STF; no LiDAR at test time.}
\label{tab:crossdomain}
\small
\setlength{\tabcolsep}{4pt}
\resizebox{\linewidth}{!}{%
\begin{tabular}{@{}lcccccccc@{}}
\toprule
 & \multicolumn{4}{c}{\textbf{O-HAZE} \cite{ancuti2018ohazedehazingbenchmarkreal} (outdoor)}
 & \multicolumn{4}{c}{\textbf{I-HAZE} \cite{DBLP:conf/acivs/AncutiATV18} (indoor)} \\
\cmidrule(lr){2-5}\cmidrule(lr){6-9}
\textbf{Method}
  & PSNR-Y\,$\uparrow$ & SSIM-Y\,$\uparrow$
  & PSNR-RGB\,$\uparrow$ & SSIM-RGB\,$\uparrow$
  & PSNR-Y\,$\uparrow$ & SSIM-Y\,$\uparrow$
  & PSNR-RGB\,$\uparrow$ & SSIM-RGB\,$\uparrow$ \\
\midrule
AOD-Net         & 15.11 & 0.5000 & 14.88 & 0.4931 & 14.69 & 0.5824 & 14.51 & 0.5724 \\
FFA-Net         & 19.20 & 0.7177 & 18.49 & 0.6968 & 15.82 & 0.6521 & 15.48 & 0.6402 \\
DehazeFormer-T  & 19.48 & 0.6839 & 18.81 & 0.6641 & 15.86 & 0.6142 & 15.56 & 0.6050 \\
DEA-Net         & 19.92 & 0.7042 & 19.20 & 0.6854 & 16.31 & 0.6455 & 16.02 & 0.6362 \\
\midrule

PhysDNet-S      & 19.89 & 0.7190 & 19.21 & 0.6989 & 16.81 & 0.6895 & 16.49 & 0.6734 \\
PhysDNet-M      & \textbf{20.45} & 0.7461 & \textbf{19.65} & 0.7253 & \textbf{16.93} & \textbf{0.6947} & \textbf{16.59} & \textbf{0.6795} \\
PhysDNet-M (NoPhy)
                & 18.92 & 0.7358 & 18.38 & 0.7174 & 16.33 & 0.6907 & 16.06 & 0.6746 \\
PhysDNet-L      & 20.31 & \textbf{0.7488} & 19.51 & \textbf{0.7279} & 16.57 & 0.6887 & 16.24 & 0.6756 \\
\bottomrule
\end{tabular}%
}
\vspace{-1em}
\end{table}

PhysDNet remains competitive without depth input. PhysDNet-M achieves the best overall performance, improving over DEA-Net by $+0.53$\,dB on O-HAZE and $+0.62$\,dB on I-HAZE, while PhysDNet-S matches or exceeds all baselines on I-HAZE. The same advantage holds on NH-HAZE (non-homogeneous outdoor haze, App.~\ref{app:nhhaze}): PhysDNet-M outperforms DEA-Net by $+0.15$\,dB PSNR-Y; removing the physics head reduces PSNR-Y by $0.54$\,dB, confirming that physics-guided training transfers despite the test distribution violating the homogeneous-scattering assumption.

The NoPhy ablation isolates the physics head: removing it reduces PSNR-Y by $1.53$\,dB on O-HAZE and $0.60$\,dB on I-HAZE. While smaller than the in-domain gap, this confirms that physics-guided training improves cross-domain generalization beyond backbone capacity.

Additional qualitative results and chrominance recovery are provided in App.~\ref{app:crossdomain}. Robustness to LiDAR sparsity is shown in App.~\ref{app:lidar_sparsity}, where PhysDNet-M retains a $+4.10$\,dB advantage at 0\% depth.

\subsection{RGB-Only Operation and Self-Supervised Colour Adaptation}
\label{sec:rgb_adaptation}

\textbf{RGB-only.} Trained from scratch on RESIDE-6K \cite{li2019benchmarking} ($D_s{=}M{=}0$, no transmission supervision), PhysDNet-M reaches 28.4\,dB on SOTS-mix (NoPhy: 28.6\,dB, consistent with \S\ref{sec:gradient}). For RGB-only deployment, the depth branch folds into constants (bit-exact), reducing compute from 8.56 to 6.33\,GMACs ($-26\%$).

\begin{figure}[t]
  \centering
  {\small
  \begin{tabular}{@{}p{0.333\linewidth}@{}p{0.333\linewidth}@{}p{0.333\linewidth}@{}}
    \centering Hazy (RTTS) &
    \centering STF zero-shot &
    \centering\arraybackslash URHI-adapted
  \end{tabular}}\\[-2pt]
  \includegraphics[width=\linewidth]{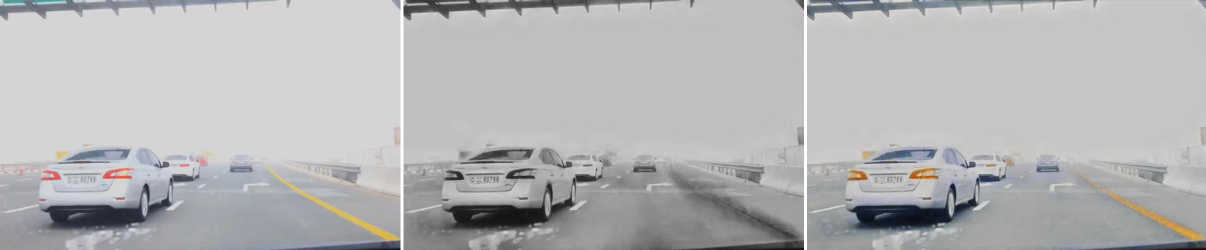}
  \vspace{-1.6em}
  \caption{\textbf{Label-free colour adaptation on real haze.} Taillights and lane markings regain colour without paired data.}
  \label{fig:rtts_adaptation}
  \vspace{-1.2em}
\end{figure}

\textbf{Learning colour without paired data.} A model trained on monochrome NIR produces near-achromatic RGB outputs. Rather than transferring chrominance from the hazy input, which carries the haze's colour cast, we learn colour through label-free adaptation on 4\,803 unpaired URHI images (5 epochs, no architecture change). The Physics Head is closed-form algebra on the coloured input, so its detached output serves as a colour teacher for the Direct Head (App.~\ref{app:color_adapt}). On 1\,297 real RTTS images \cite{li2019benchmarking}, colourfulness \cite{hasler2003colorfulness} rises from 0.03 to 12.8 with the best no-reference scores of all conditions (NIQE 4.74, BRISQUE 33.4 vs.\ 5.35/38.3 for the hazy input; Fig.~\ref{fig:rtts_adaptation}). Adaptation specializes the model to the target domain ($-5.1$\,dB on O-HAZE), so Table~\ref{tab:crossdomain} reports the unadapted model.

\subsection{Real-World Fog Validation}
\label{sec:real_world}

Since STF real-fog frames lack ground truth, we evaluate via
\textbf{dual-head consistency}, i.e., agreement between
$\hat{J}_{\text{dir}}$ and $\hat{J}_{\text{phys}}$
(Table~\ref{tab:consistency}).

\begin{table}
\centering
\caption{%
  \textbf{Dual-head consistency on real STF fog.}
  High PSNR and correlation indicate strong agreement.
}
\label{tab:consistency}
\small
\begin{tabular}{@{}lccc@{}}
\toprule
\textbf{Metric}
  & \textbf{Dense Fog}
  & \textbf{Light Fog}
  & \textbf{Interpretation} \\
\midrule
MSE                     & 0.0002 & 0.0006 & Very low gap \\
PSNR (dB) $\uparrow$   & 37.48  & 33.83  & High agreement \\
Pearson $r$ $\uparrow$  & 0.9947 & 0.9912 & Near-perfect match \\
\bottomrule
\end{tabular}
\end{table}

Both heads show near-perfect agreement (PSNR $>$30\,dB,
$r>0.99$), confirming that the physics inversion produces outputs
consistent with the direct head. Consistency is higher in dense fog
(37.48 vs.\ 33.83\,dB), aligning with the
density-adaptive gradient amplification (\S\ref{sec:gradient}).

We further validate real-haze performance using no-reference metrics
(App.~\ref{app:niqe}): PhysDNet-M achieves the best NIQE on O-HAZE
(3.92 vs.\ 4.64 for DEA-Net), confirming transfer beyond synthetic data. Across all official STF weather splits, including rain and snow, agreement remains high and is again highest in dense fog ($r{\geq}0.993$ throughout; App.~\ref{app:weather}).

\subsection{Edge Deployment}
\label{sec:edge_deployment}
PhysDNet's restriction to NPU-friendly operators avoids softmax-attention, LayerNorm, and deformable convolutions, which block baseline compilation.

We validate deployability on two representative edge targets:
(i) a fixed-function 26\,TOPS NPU ($<$4\,W TDP), representative of
strict automotive perception SoCs, and
(ii) the NVIDIA Jetson Orin Nano Super (8\,GB), representative of
more permissive ADAS compute platforms. All models are exported to ONNX from the same trained checkpoints and evaluated at
$256{\times}256$. Table~\ref{tab:edge-merged} reports compilation
status, throughput, and power.

\begin{table}
\centering
\caption{%
  \textbf{Unified edge-deployment benchmark}
  ($256{\times}256$, batch 1, 50 warmup + 300 timed iterations on the Jetson Orin Nano). Fixed-function NPU uses INT8 post-training quantization; Jetson uses TensorRT FP16. \xmark{} indicates compilation failure due to unsupported operators.
  Full CUDA FP32 numbers and power profiles in
  Appendix~\ref{app:edge_full}.}
\label{tab:edge-merged}
\small
\begin{tabular}{@{}lrrcccc@{}}
\toprule
\multirow{2}{*}{\textbf{Model}} &
\multirow{2}{*}{\textbf{Params\,(M)}} &
\multirow{2}{*}{\textbf{GMACs}} &
\multicolumn{2}{c}{\textbf{Fixed-function NPU}} &
\multicolumn{2}{c}{\textbf{Jetson Orin Nano}} \\
\cmidrule(lr){4-5}\cmidrule(lr){6-7}
 & & & Compile & FPS & FPS & $\overline{P}$\,(W) \\
\midrule
AOD-Net LiDAR        & 0.002 &  0.06 & \cmark & \textbf{1220}         & 327.0          &  4.94 \\
FFA-Net LiDAR        & 0.77  & 24.68 & \xmark & ---         & 46.9           &  8.21 \\
DehazeFormer-T LiDAR & 1.36  &  2.73 & \xmark & ---         & 47.1           & 12.38 \\
DEA-Net LiDAR        & 3.65  & 16.44 & \xmark & ---         & 28.5           & 10.17 \\
\midrule
PhysDNet-S (ours)    & 0.72  &  2.24 & \cmark & \textbf{99} & \textbf{108.3} & 13.45 \\
PhysDNet-M (ours)    & 2.95  &  8.56 & \cmark & 45          & 58.5           & 12.38 \\
PhysDNet-L (ours)    & 6.59  & 18.98 & \cmark & 29          & 38.8           & 11.80 \\
\bottomrule
\end{tabular}
\vspace{-1.5em}
\end{table}

\begin{figure}[t]
  \centering
  \includegraphics[width=0.95\linewidth]{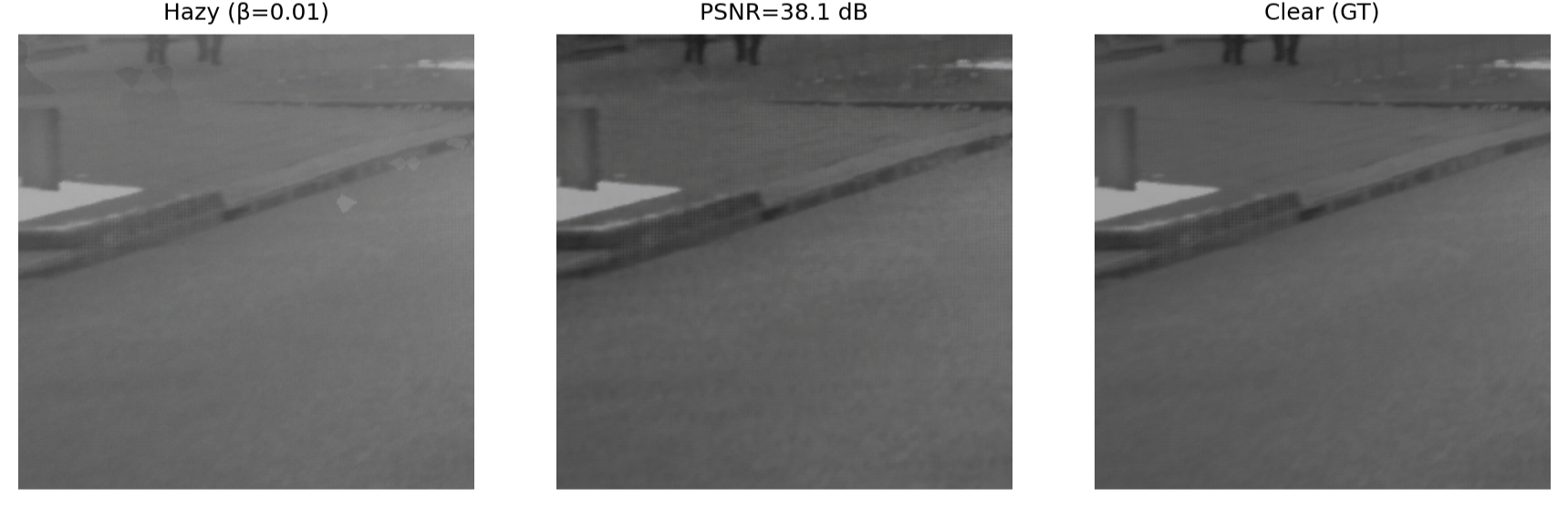}
  \vspace{-0.5em}
  \caption{%
    \textbf{INT8 NPU inference on a representative STF sample} ($\beta{=}0.01$).
    Hazy input, PhysDNet-M restored on the fixed-function NPU (38.1\,dB), and ground truth. INT8 deployment preserves restoration quality within the training range; full $\beta$-sweep in App.~\ref{app:edge_quant}.}
  \label{fig:hailo_sample}
  \vspace{-1em}
\end{figure}

\textbf{Fixed-function NPU.}
All models are compiled via vanilla INT8 post-training quantization (64-sample calibration, no QAT). The ${\sim}1.5$\,dB PSNR drop (App.~\ref{app:edge_quant}) is a conservative lower bound on quantization robustness. \cref{fig:hailo_sample} shows that within the training range, the INT8 output is visually indistinguishable from the FP32 reference.

\textbf{Under our evaluation protocol (standard ONNX export and vendor compiler with INT8 post-training quantization), only PhysDNet and AOD-Net successfully compile}, but AOD-Net's restoration quality is insufficient for deployment
(20.95\,dB, \S\ref{sec:base}). All other baselines fail due to unsupported
operators (e.g., attention, LayerNorm). PhysDNet is thus the only method
that combines competitive accuracy with NPU compatibility, while exceeding
the 10\,Hz LiDAR rate.

\textbf{Jetson Orin Nano.}
All primary baselines compile under CUDA+TensorRT. PhysDNet-M runs $1.25\times$ faster than FFA-Net and $2.05\times$ faster than DEA-Net; PhysDNet-L remains $1.36\times$ faster than DEA-Net. AOD-Net's 327\,FPS lacks deployable quality (\S\ref{sec:base}).

\section{Related Work}\label{sec:related_work}

\paragraph{Physics-Guided Dehazing}
Classical dehazing combines the atmospheric scattering model with
image priors such as the Dark Channel Prior~\cite{DBLP:conf/cvpr/HeST09}.
Learned physics-guided methods incorporate atmospheric structure
through model-based parameterizations or consistency
objectives~\cite{DBLP:conf/iccv/LiPWXF17,PEDNet2024,PHPDeNet2025}.
PhysDNet emphasizes the optimization effect of differentiating
through a fixed inversion: its transmission-dependent Jacobian
shapes supervision in the shared backbone. The Direct Head remains
unconstrained by the inversion at inference, allowing residual
corrections beyond the assumed image-formation model. This separates
the role of physics during training from the form of the deployed
restoration output.

\paragraph{Image-Only and Depth-Guided Dehazing}
Image-only CNN and transformer methods infer restoration cues from
appearance~\cite{DBLP:conf/aaai/QinWBXJ20,Song_2023,DBLP:journals/tip/ChenHL24}.
Depth-aware approaches additionally exploit binocular disparity,
monocular depth estimation, or LiDAR-derived
geometry~\cite{DBLP:conf/cvpr/PangNXHL20,DBLP:journals/pr/LiangLCWLL25,DBLP:journals/sensors/ChungKK22,DBLP:journals/corr/Tran2022}.
PhysDNet directly combines gated NIR imagery with sparse LiDAR
covering approximately 3\% of image pixels, using mask-aware
multi-scale gating without a depth-completion stage. LiDAR is treated
as a weather-attenuated geometric prior: we assume retained
strongest returns provide useful range measurements, while missing
returns are represented explicitly by a validity mask. Neither
complete coverage nor immunity to fog is assumed. The zero-depth
experiments in Appendix~\ref{app:lidar_sparsity} quantify operation
when this prior is unavailable.

\paragraph{Edge Deployment and Baseline Selection}
We target single-pass restoration within 33\,ms and 15\,W, compatible with fixed-function INT8 accelerators~\cite{dhahri2025quanttrimpracticeimprovedcrossplatform}, which motivates our comparison with compact, toolchain-supported architectures. Recent heavyweight~\cite{uvm,DFFNet,ksformer} and diffusion-based~\cite{PhysGuidedDiffusion2025,FrDiff2025,yu2023dehazeddpm,xiong2025rshazediff,lin2025hazediff,luo2023irsde} methods trade compute for quality; we measure rather than assume their cost (App.~\ref{app:model_selection}). The fastest, Ksformer~\cite{ksformer}, reaches 2.8\,FPS in TensorRT FP16, while DehazeDDPM~\cite{yu2023dehazeddpm} and RSHazeDiff~\cite{xiong2025rshazediff} need an estimated 3.1 and 17.0\,s per image. PhysDNet confines the inversion to training, adding no cost to the deployed Direct Head.

\section{Conclusion}

We introduced PhysDNet, a lightweight multimodal dehazing architecture that uses a physics-constrained dual-head objective to shape optimization dynamics. Differentiating through a fixed Koschmieder inversion induces density-adaptive gradient amplification, concentrating learning on severely degraded regions at zero inference cost. This mechanism yields state-of-the-art performance on Seeing Through Fog, consistent gains under cross-domain evaluation, and robustness to LiDAR sparsity. PhysDNet is also the only evaluated method that combines strong restoration quality with fixed-function low-power NPU deployment, achieving 29-99\,FPS at $<$4\,W under INT8 quantization. 

Beyond dehazing, our results suggest that physical models can serve not only as priors on outputs, but as mechanisms for shaping the learning signal itself, offering a new paradigm for efficient and deployable perception under physical constraints.

\textbf{Limitations.}
PhysDNet is trained on synthetic Koschmieder haze; NoPhy ablations on three real-fog benchmarks (O-HAZE, I-HAZE, NH-HAZE) confirm transfer, including under non-homogeneous fog (App.~\ref{app:nhhaze}). Broader cross-platform multi-modal validation beyond STF remains open. Without color supervision, the model produces near-achromatic RGB outputs requiring heuristic chrominance recovery (\S\ref{sec:ohaze}).

\section*{Acknowledgments}
This work was partially supported by the Ministry of Education and Science of Bulgaria, as part of the Bulgarian National Roadmap for Research Infrastructure (support for INSAIT) and the European Union’s Horizon Europe – the Framework Programme for Research and Innovation, under grant agreement 101168521.\\
Rayen Dhahri's research was self-funded, including all computational resources used for this work.

\bibliographystyle{unsrtnat}
\bibliography{ref}

\newpage
\appendix

\section{Broader impact}\label{app:broader_impact}
PhysDNet improves the ability of resource-constrained edge systems to perceive in adverse weather, with direct applications to autonomous driving, robotics, and ADAS. By recovering semantically meaningful structure and improving downstream object detection in dense fog, it may help reduce weather-related accidents and improve road safety. Its design also supports \emph{Green AI} by enabling high-quality, physics-grounded restoration on ultra-low-power fixed-function NPUs ($<\textbf{4\,W}$), reducing reliance on energy-intensive GPU inference.

At the same time, this technology has dual-use potential and could support unconsented surveillance or military deployment in conditions where poor visibility would otherwise provide cover. It also introduces safety risks in critical systems if exposed to out-of-distribution weather, such as heterogeneous smoke or severe occlusions that violate the assumed Koschmieder model. For this reason, PhysDNet should be used only as part of a  sensor-fusion stack, together with uncertainty estimation and hardware fail-safes.

\section{Extended Background}
\label{app:ext_background}

\paragraph{Ill-posed nature of monocular dehazing.}
Recovering $J(x)$ from a single $I(x)$ is fundamentally ill-posed: both $t(x)$ and $A$ are unknown, and hand-crafted priors such as the Dark Channel Prior~\cite{he2010single} frequently collapse under non-homogeneous real-world fog.

\paragraph{Active sensing and gated imaging mechanics.}
Modern autonomous platforms combine sparse LiDAR, which provides weather-invariant depth measurements $d(x)$, with active time-gated NIR imaging~\cite{DBLP:journals/corr/abs-1912-02889}. By delaying sensor exposure after emitting a synchronised NIR laser pulse, a gated camera physically rejects nearby backscatter and integrates only photons from a target depth slice, drastically improving the signal-to-noise ratio in dense fog. This gated imagery combined with sparse LiDAR forms the physical foundation of PhysDNet's multi-modal input.

\section{Baseline Selection Rationale}
\label{app:model_selection}

\Cref{tab:model_rationale} measures every candidate under our edge protocol (Jetson Orin Nano Super, TensorRT FP16 where export succeeds). The primary baselines meet the single-pass deployment criterion. The fastest recent heavyweight method that runs, Ksformer, reaches 2.8\,FPS, UVM-Net cannot build an engine, and the diffusion methods need 3.1--17\,s per image. Even PhysDNet-L runs at 38.8\,FPS within the latency and power targets.
\begin{table}
\centering
\caption{%
\textbf{Measured Edge Deployment and Baseline Selection.}
Compute is profiled at $256{\times}256$, batch~1.
Jetson results report latency, throughput, and mean module power.
Successful exported models use TensorRT FP16; DFFNet uses
PyTorch eager FP32.
Diffusion per-image costs are derived from measured network-call
latency and the stated number of network evaluations.
Unmeasured related methods are listed separately.
}
\label{tab:model_rationale}
\setlength{\tabcolsep}{4pt}
\renewcommand{\arraystretch}{1.15}
\resizebox{\linewidth}{!}{%
\begin{tabular}{@{}lrrrrr p{4.5cm}@{}}
\toprule
\textbf{Method} &
\textbf{\shortstack{Params\\(M)}} &
\textbf{\shortstack{Compute\\(GMACs/image)}} &
\textbf{\shortstack{Jetson latency\\(ms/image)}} &
\textbf{FPS} &
\textbf{\shortstack{Power\\(W)}} &
\textbf{Explanation} \\
\midrule

\multicolumn{7}{@{}l}{%
\cellcolor{rowA}\textbf{Primary Comparison Baselines}} \\

DCP~\cite{DBLP:conf/cvpr/HeST09}
& 0 & --- & --- & --- & ---
& Classical physical-prior baseline. \\

AOD-Net~\cite{DBLP:conf/iccv/LiPWXF17}
& 0.002 & 0.06 & 3.1 & 327.0 & 4.94
& Ultra-lightweight model-based CNN. \\

DehazeFormer-T~\cite{Song_2023}
& 1.36 & 2.73 & 21.2 & 47.1 & 12.38
& Compact transformer baseline. \\

DEA-Net~\cite{DBLP:journals/tip/ChenHL24}
& 3.65 & 16.44 & 35.1 & 28.5 & 10.17
& Competitive CNN baseline; exceeds the latency target. \\

FFA-Net~\cite{DBLP:conf/aaai/QinWBXJ20}
& 0.77 & 24.68 & 21.3 & 46.9 & 8.21
& Feature-fusion attention baseline. \\

\midrule
\multicolumn{7}{@{}l}{%
\cellcolor{physBg}\textbf{Measured Heavyweight Baselines}} \\

UVM-Net~\cite{uvm}$^\dagger$
& 1003.9 & 250.9 & --- & --- & ---
& TensorRT engine construction fails: out-of-memory
  in the tested implementation. \\

DFFNet~\cite{DFFNet}$^\dagger$
& 17.8 & 61.3 & 157.0 & 6.4 & ---
& ONNX export fails on FFT operations;
  latency measured in eager FP32. \\

Ksformer~\cite{ksformer}$^\dagger$
& 5.5 & 109.0 & 360.0 & 2.8 & 19.8
& TensorRT FP16; exceeds both latency and power targets. \\

\midrule
\multicolumn{7}{@{}l}{%
\cellcolor{physBg}\textbf{Measured Diffusion Backbones and Sampling Costs}} \\

DehazeDDPM~\cite{yu2023dehazeddpm}$^\ddagger$
& 97.8 & 4\,509 & $\approx$3\,100 & 0.32 & 20.6
& 50 network evaluations per image. \\

RSHazeDiff~\cite{xiong2025rshazediff}$^\ddagger$
& 109.7 & 13\,523 & $\approx$17\,000 & 0.06 & 16.0
& 10 steps across 169 overlapping patches:
  1,690 network evaluations per image. \\

\midrule
\multicolumn{7}{@{}l}{%
\cellcolor{rowC}\textbf{PhysDNet Deployment Reference}} \\

\textbf{PhysDNet-L}
& 6.59 & 18.98 & \textbf{25.8} & \textbf{38.8} & 11.80
& TensorRT FP16; Even our largest variant satisfies the stated Jetson
  latency and power targets. \\

\midrule
\multicolumn{7}{@{}l}{%
\cellcolor{auxBg}\textbf{Other Related Methods: Qualitative Positioning}} \\

HazeDiff~\cite{lin2025hazediff}
& --- & --- & --- & --- & ---
& \multirow{3}{4.5cm}{%
Iterative generative restoration approaches outside the
primary single-pass comparison. Retained for related-work
positioning; not included in the measured benchmark above.} \\

IR-SDE~\cite{luo2023irsde}
& --- & --- & --- & --- & ---
& \\

FrDiff~\cite{FrDiff2025}
& --- & --- & --- & --- & ---
& \\[5.5ex]

\bottomrule
\end{tabular}%
}

\vspace{3pt}
\begin{minipage}{\linewidth}
\footnotesize
\textbf{Protocol.}
Jetson Orin Nano Super, MAXN, batch~1;
engine construction excluded from inference timing;
mean module power recorded with \texttt{tegrastats}.
Primary learned-baseline rows use the LiDAR-augmented
configurations from Table~\ref{tab:edge-merged}.
Their displayed latency is calculated as $1000/\mathrm{FPS}$.

\smallskip
$^\dagger$DFFNet and Ksformer are reimplementations matched
to published architecture descriptions and parameter budgets;
their random weights support compute and latency profiling,
not restoration-quality comparisons.
UVM-Net uses an official-code port with a pure-PyTorch
selective-scan fallback; its failure applies to this tested path.

\smallskip
$^\ddagger$DehazeDDPM: 90.2\,GMACs and 62.7\,ms per call,
using a 50-call benchmark setting.
RSHazeDiff: 8.0\,GMACs and approximately 10.0\,ms per call,
using $10\times169$ patch evaluations.
Per-image compute and latency multiply per-call measurements
by these evaluation counts; latency excludes sampler and
patch-aggregation overhead. Totals use unrounded measurements.
Dashes indicate unavailable or unmeasured results.
\end{minipage}
\end{table}

\section{Architecture Details}
\label{app:arch_details}

Each encoder stage consists of two DWSep blocks with a residual shortcut, followed by $2{\times}$ max-pooling; the four stages produce feature maps at $H$, $H/2$, $H/4$, and $H/8$ (bottleneck). The image branch accepts a 3-channel input (monochrome NIR replicated or colour); the depth branch accepts 2 channels (sparse depth $D_s$ and binary validity mask $M$). A Squeeze-and-Excitation (SE) channel attention block~\cite{DBLP:journals/corr/abs-1709-01507} with reduction ratio 4 is applied once at the bottleneck after gated fusion, recalibrating channel responses before decoding. The decoder consists of three upsampling stages, each performing bilinear $2\times$ upsampling followed by channel-wise concatenation with the corresponding fused skip features and refinement through a DWSep residual block. The bottleneck fused features ($8c$ channels at $H/8$) serve as the decoder input; skip connections from scales 3, 2, and 1 progressively reintroduce spatial detail. The head projections use $1{\times}1$ convolutions throughout; airlight is predicted via global average pooling of the bottleneck followed by a 3-unit fully-connected layer.

\section{Training Details and Hyperparameters}
\label{app:hyperparams}
 
All PhysDNet variants share a common training recipe and loss weights, differing only in base channel width $c$. The \textbf{L} variant additionally enables mixed-precision (AMP) to fit within VRAM at $c{=}96$.

\begin{table}[H]
\centering
\caption{%
  \textbf{Training recipe.}
  Shared across all variants except where noted.
  $^\dagger$\,AMP enabled for PhysDNet-L only.
}
\label{tab:recipe}
\small
\begin{tabular}{@{} l c c c @{}}

\hdr{Hyperparameter} & \hdr{PhysDNet-S} & \hdr{PhysDNet-M} & \hdr{PhysDNet-L} \\

\rowcolor{rowA}
Base channel width $c$            & 32   & 64   & 96   \\
SE attention                      & \checkmark & \checkmark & \checkmark \\
\rowcolor{rowA}
Epochs                            & 500  & 500  & 500  \\
Batch size                        & 16   & 16   & 8$^\dagger$ \\
\rowcolor{rowA}
Optimizer                         & \multicolumn{3}{>{\columncolor{rowA}}c}{AdamW} \\
Learning rate                     & \multicolumn{3}{c}{$2 \times 10^{-4}$} \\
\rowcolor{rowA}
Weight decay                      & \multicolumn{3}{>{\columncolor{rowA}}c}{$1 \times 10^{-4}$} \\
LR scheduler                      & \multicolumn{3}{c}{Cosine annealing $\to 10^{-6}$} \\
\rowcolor{rowA}
Warmup epochs                     & \multicolumn{3}{>{\columncolor{rowA}}c}{10} \\
Gradient clipping                 & \multicolumn{3}{c}{$\|\nabla\|_2 \leq 1.0$} \\
\rowcolor{rowA}
Crop size                         & \multicolumn{3}{>{\columncolor{rowA}}c}{$256 \times 256$} \\
Data augmentation                 & \multicolumn{3}{c}{Random flip + crop} \\
\rowcolor{rowA}
Mixed precision (AMP)             & \texttimes & \texttimes & \checkmark \\
\midrule
\rowcolor{rowB}
$\beta$ range (haze synthesis)     & \multicolumn{3}{>{\columncolor{rowB}}c}{$\mathcal{U}(0.005,\, 0.04)$} \\
\rowcolor{rowB}
$A_k$ range (airlight per channel) & \multicolumn{3}{>{\columncolor{rowB}}c}{$\mathcal{U}(0.7,\, 1.0)$} \\
\rowcolor{rowB}
Max LiDAR depth                    & \multicolumn{3}{>{\columncolor{rowB}}c}{120\,m} \\
\bottomrule
\end{tabular}
\end{table}
 
\begin{table}[H]
\centering
\caption{
  \textbf{Loss component weights ($\lambda$).}
  All variants share the same loss weights and contrastive activation 
  schedule (epoch 185); they differ only in base channel width $c$.
  \colorbox{dataBg}{\strut\,Data-driven\,}\,/\,%
  \colorbox{physBg}{\strut\,Physics\,}\,/\,%
  \colorbox{auxBg}{\strut\,Auxiliary\,} shading indicates loss family.
}
\label{tab:loss_weights}
\small

\begin{tabular}{@{} l l l c c c l @{}}
\hdr{Loss} & \hdr{Symbol} & \hdr{Supervises}
  & \hdr{S} & \hdr{M} & \hdr{L} & \hdr{Role} \\
\rowcolor{dataBg}
Reconstruction
  & $\mathcal{L}_{\text{rec}}$
  & $\hat{J}_{\text{dir}}$ vs.\ $J_{\text{gt}}$
  & 1.0 & 1.0 & 1.0
  & Primary restoration \\
\rowcolor{dataBg}
Perceptual
  & $\mathcal{L}_{\text{perc}}$
  & VGG-16 features
  & 0.05 & 0.05 & 0.05
  & Feature-space similarity \\
\rowcolor{dataBg}
Gradient
  & $\mathcal{L}_{\text{grad}}$
  & $\nabla\hat{J}_{\text{dir}}$ vs.\ $\nabla J_{\text{gt}}$
  & 0.5 & 0.5 & 0.5
  & Edge preservation \\
\midrule
\rowcolor{physBg}
Physics recon.
  & $\mathcal{L}_{\text{pr}}$
  & $\hat{J}_{\text{phys}}$ vs.\ $J_{\text{gt}}$
  & 1.0 & 1.0 & 1.0
  & Physics-head fidelity \\
\rowcolor{physBg}
Transmission
  & $\mathcal{L}_{t}$
  & $\hat{t}$ vs.\ $t_{\text{gt}}$
  & 2.0 & 2.0 & 2.0
  & LiDAR-supervised $t$ \\
\rowcolor{physBg}
Smoothness
  & $\mathcal{L}_{\text{sm}}$
  & $\nabla\hat{t}$
  & 0.1 & 0.1 & 0.1
  & Edge-aware regularization \\
\rowcolor{physBg}
Consistency
  & $\mathcal{L}_{\text{con}}$
  & $\hat{J}_{\text{dir}}$ vs.\ $\hat{J}_{\text{phys}}$
  & 0.2 & 0.2 & 0.2
  & Dual-head agreement \\
\midrule
\rowcolor{auxBg}
FFT
  & $\mathcal{L}_{\text{fft}}$
  & Frequency domain
  & 0.1 & 0.1 & 0.1
  & High-frequency detail \\
\rowcolor{auxBg}
Contrastive
  & $\mathcal{L}_{\text{contra}}$
  & Latent features
  & 0.1 & 0.1 & 0.1
  & Physics-aware bound \\
\rowcolor{auxBg}
start epoch
  &
  &
  & 185 & 185 & 185
  & Delayed activation \\

\end{tabular}
\end{table}
 
\begin{table}[H]
\centering
\caption{%
  \textbf{PhysDNet model family.}
  Scaling the base channel width $c$ yields three variants spanning the
  accuracy--efficiency Pareto front.
  All variants share the same dual-encoder, gated-fusion, SE-decoder
  topology (\S\ref{sec:architecture}) and dual-head design
  (\S\ref{sec:heads}).
  Physics head parameters are excluded from the inference count as
  only the direct head is deployed.
}
\label{tab:variants}
\small
\begin{tabular}{@{} l c c c c c c @{}}

\hdr{Variant}
  & \hdr{$c$}
  & \hdr{SE Attn.}
  & \hdr{Params (M)}
  & \hdr{GMACs}
  & \hdr{Bottleneck}
  & \hdr{Target} \\

\rowcolor{rowC}
PhysDNet-S & 32 & \checkmark & 0.72 &  2.24 & $256 \times \tfrac{H}{8} \times \tfrac{W}{8}$ & Ultra-low-power NPU \\
PhysDNet-M & 64 & \checkmark & 2.95 &  8.56 & $512 \times \tfrac{H}{8} \times \tfrac{W}{8}$ & Edge GPU / mid-range NPU \\
\rowcolor{rowC}
PhysDNet-L & 96 & \checkmark & 6.59 & 18.98 & $768 \times \tfrac{H}{8} \times \tfrac{W}{8}$ & High-perf.\ edge accelerator \\
\end{tabular}
\end{table}

\paragraph{Convergence and Training Budgets.}
All models were trained until performance plateaued under their respective training configurations, subject to a fixed computational budget. In practice, we observe that PSNR improvements saturate after approximately 80 epochs for most baselines, with additional training yielding only marginal gains (typically $<0.3$\,dB). This behavior is consistent across all evaluated methods, indicating that extended training primarily increases compute cost without significantly affecting relative performance rankings. PhysDNet variants were trained for 500 epochs to ensure the complex multi-term objective (including late-activated contrastive losses) fully converged, though primary restoration metrics similarly plateaued early in the schedule. Experiments require 24 GB VRAM GPU.

\section{Gradient Weighting under the Scattering Model: Full Derivation}
\label{sec:theory_proof}

Section~\ref{sec:gradient} empirically characterised the $(\hat{t}+\epsilon)^{-2}$ gradient structure induced by differentiating the Koschmieder inversion and briefly noted its structural correspondence to the inversion's noise variance. This appendix provides the formal derivations of this variance-gradient relationship, alongside the airlight-gradient analysis, the bounded-gradient discussion, and a scope statement delineating what the analysis does and does not claim.

\paragraph{Setup.}
We adopt the standard atmospheric scattering model with additive sensor noise~\cite{DBLP:conf/cvpr/HeST09,s22114169,bianco2015}:
\begin{equation}
    I(x) = J(x)\,t(x) + A\bigl(1-t(x)\bigr) + \xi(x), \qquad \xi(x) \sim \mathcal{N}(0,\sigma^2),
\label{eq:noisy_asm_app}
\end{equation}
with $\xi(x)$ independent across pixels. Given estimates $\hat{t}(x),\hat{A}$, the physics head computes the $\epsilon$-stabilised inversion
\begin{equation}
    \hat{J}_{\mathrm{phys}}(x) \;=\; \frac{I(x) - \hat{A}\bigl(1-\hat{t}(x)\bigr)}{\hat{t}(x)+\epsilon},
\label{eq:phys_inv_eps_app}
\end{equation}
consistent with Eq.~\eqref{eq:physics_inversion} in the main text.

\paragraph{Proposition 1 (Transmission-Scaled Noise Variance).}
\textit{Under Eq.~\eqref{eq:noisy_asm_app} and first-order estimates $\hat{t}(x) = t(x) + \delta_t$, $\hat{A} = A + \delta_A$ with $|\delta_t|,|\delta_A| = O(\delta)$ and $\delta \ll t_{\min}$, the inverted radiance admits}
\begin{equation}
    \hat{J}_{\mathrm{phys}}(x) \;=\; J(x) + \frac{\xi(x)}{t(x)} + O(\delta),
\qquad
    \operatorname{Var}\bigl[\hat{J}_{\mathrm{phys}}(x)\bigr] \;=\; \frac{\sigma^2}{t(x)^2} + O(\delta).
\label{eq:variance_scaling_app}
\end{equation}

\textit{Proof.} Substituting Eq.~\eqref{eq:noisy_asm_app} into Eq.~\eqref{eq:phys_inv_eps_app}:
\[
\hat{J}_{\mathrm{phys}} \;=\; \frac{Jt + A(1-t) + \xi - (A+\delta_A)\bigl(1-(t+\delta_t)\bigr)}{(t+\delta_t)+\epsilon}.
\]
Expanding the bracket in the numerator gives
$(A+\delta_A)(1-t-\delta_t) = A(1-t) - A\delta_t + \delta_A(1-t) - \delta_A\delta_t$,
so the $A(1-t)$ terms cancel and
\[
\text{numerator} \;=\; Jt + \xi + A\delta_t - \delta_A(1-t) + O(\delta^2).
\]
For the denominator, since $\delta,\epsilon \ll t$, we have the first-order expansion
\[
\frac{1}{t+\delta_t+\epsilon}
\;=\; \frac{1}{t}\cdot\frac{1}{1+(\delta_t+\epsilon)/t}
\;=\; \frac{1}{t} - \frac{\delta_t+\epsilon}{t^2} + O(\delta^2,\epsilon^2).
\]
Multiplying numerator and denominator and retaining terms up to $O(\delta)$:
\begin{align*}
\hat{J}_{\mathrm{phys}}
&\;=\; \bigl[Jt + \xi + A\delta_t - \delta_A(1-t)\bigr]\cdot\tfrac{1}{t}
   \;-\; Jt\cdot\tfrac{\delta_t+\epsilon}{t^2}
   \;+\; O(\delta^2,\epsilon) \\[2pt]
&\;=\; J + \tfrac{\xi}{t} + \tfrac{A\delta_t - \delta_A(1-t)}{t}
   \;-\; \tfrac{J\delta_t}{t} + O(\delta^2,\epsilon) \\[2pt]
&\;=\; J + \tfrac{\xi}{t} - \tfrac{(J-A)\delta_t + (1-t)\delta_A}{t} + O(\delta^2,\epsilon),
\end{align*}
where the cross-terms (e.g.\ $\xi\cdot O(\delta,\epsilon)$, $A\delta_t\cdot O(\delta,\epsilon)$) all collapse into $O(\delta^2,\epsilon)$.
The deterministic perturbation terms are $O(\delta)$; only $\xi/t$ contributes stochastic variance. Since $\xi\sim\mathcal{N}(0,\sigma^2)$,
$\operatorname{Var}[\hat{J}_{\mathrm{phys}}] = \sigma^2/t^2 + O(\delta)$.
\hfill$\square$

\begin{figure}[H]\centering
\includegraphics[width=0.8\linewidth]{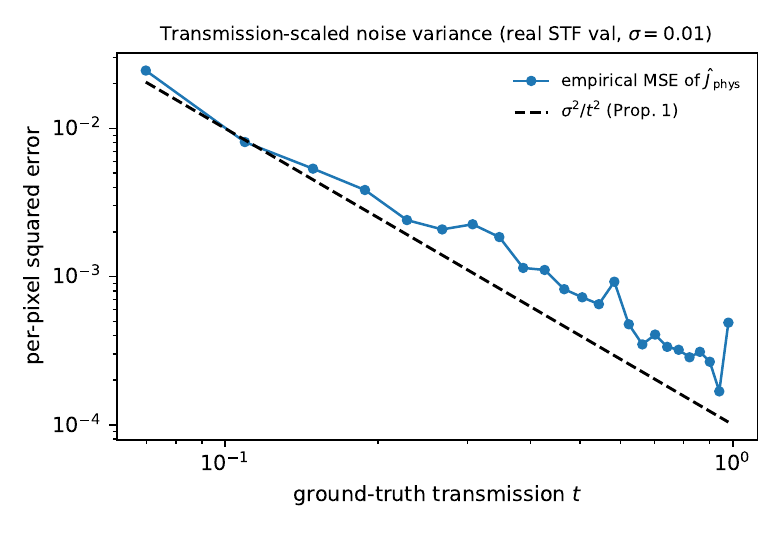}
\caption{\textbf{Empirical test of Proposition~1.} Binned squared error of $\hat J_{\text{phys}}$ vs.\ ground-truth transmission, overlaid with $\sigma^2/t^2$.}
\label{fig:prop1}
\end{figure}

\paragraph{Quantitative validation of Proposition~1.}
\label{app:prop1_validation}
On 300 STF validation frames (clear day/night splits) with synthetic haze and Gaussian sensor noise ($\sigma{\approx}0.01$), we bin the per-pixel squared error of the parameter-free inversion of the released PhysDNet-M by ground-truth transmission (24 bins, clamp-saturated pixels excluded). The empirical curve follows $\sigma^2/t^2$ across two orders of magnitude (log-log $r{=}0.981$; Fig.~\ref{fig:prop1}); the small uniform offset is consistent with the $O(\delta)$ estimation term. This is an MSE trend consistent with the predicted variance law, not a direct variance measurement on natural fog. It also identifies the low-transmission noise visible in Physics Head panels (Fig.~\ref{fig:qualitative_results}) as the predicted structure, which the Direct Head does not inherit.

\paragraph{Proposition 2 (Structured Jacobian Scaling).}
The Jacobian of the inversion with respect to transmission is
\begin{equation}
    \frac{\partial \hat{J}_{\mathrm{phys}}}{\partial \hat{t}}(x)
    = \frac{\hat{A} - I(x)}{(\hat{t}(x)+\epsilon)^2},
\label{eq:jacobian_scaling_app}
\end{equation}
treating $(I,\hat{A})$ as fixed with respect to $\hat{t}$.

\textit{Derivation.}
Write $\hat{J}_{\mathrm{phys}} = u/v$ with
$u(\hat{t}) = I - \hat{A}(1-\hat{t}) = I - \hat{A} + \hat{A}\hat{t}$
and $v(\hat{t}) = \hat{t}+\epsilon$.
Then $\partial u/\partial\hat{t} = \hat{A}$, $\partial v/\partial\hat{t} = 1$, and the quotient rule yields
\[
\frac{\partial\hat{J}_{\mathrm{phys}}}{\partial\hat{t}}
\;=\; \frac{\hat{A}\,(\hat{t}+\epsilon) - \bigl[I - \hat{A}(1-\hat{t})\bigr]}{(\hat{t}+\epsilon)^2}
\;=\; \frac{\hat{A}-I + \hat{A}\,\epsilon}{(\hat{t}+\epsilon)^2}
\;=\; \frac{\hat{A}-I}{(\hat{t}+\epsilon)^2} + O(\epsilon),
\]
which gives Eq.~\eqref{eq:jacobian_scaling_app} after dropping the $O(\epsilon)$ term consistent with the main text.
\hfill$\square$

This expression reveals a \emph{structural} inverse-square dependence on transmission through the denominator $(\hat{t}+\epsilon)^{-2}$. Consequently, the computational graph intrinsically amplifies gradients in low-transmission (dense fog) regions.
The effective gradient magnitude additionally depends on the residual term $(\hat{A} - I(x))$, which introduces a data-dependent modulation. Under the scattering model, $(\hat{A} - I(x)) \approx t(x)(A - J(x))$ when estimates are accurate, partially attenuating the scaling in well-reconstructed regions. Nevertheless, the dominant $(\hat{t}+\epsilon)^{-2}$ structure ensures that low-transmission pixels receive substantially larger gradient magnitudes throughout training.
Importantly, this amplification arises directly from differentiating the physical inversion and does not rely on explicit weighting heuristics, auxiliary parameters, or curriculum schedules.

\paragraph{Corollary (Loss-gradient scaling).}
Under $\mathcal{L}_{\mathrm{pr}} = \|\hat{J}_{\mathrm{phys}} - J_{\mathrm{gt}}\|_1$,
\begin{equation}
    \frac{\partial \mathcal{L}_{\mathrm{pr}}}{\partial \hat{t}(x)} \;=\; \operatorname{sign}\!\bigl(\hat{J}_{\mathrm{phys}}(x) - J_{\mathrm{gt}}(x)\bigr)\cdot\frac{\hat{A}-I(x)}{(\hat{t}(x)+\epsilon)^2},
\label{eq:loss_grad_app}
\end{equation}
and under an $\ell_2$ loss the $\mathrm{sign}$ is replaced by $2\bigl(\hat{J}_{\mathrm{phys}}-J_{\mathrm{gt}}\bigr)$. In both cases the amplification propagates unchanged into the encoder weights via $\partial\hat{t}/\partial\theta_{\mathrm{enc}}$.

\paragraph{Airlight gradient: physically correct supervision routing.}
The magnitude of the airlight gradient is $|\partial\hat{J}_{\mathrm{phys}}/\partial\hat{A}| = |1-\hat{t}|/(\hat{t}+\epsilon)$, which scales as $\hat{t}^{-1}$ for small $\hat{t}$ and vanishes as $\hat{t}\to 1$. The network therefore learns to estimate airlight primarily from foggy regions, where atmospheric light dominates the observation, and ignores clear-air pixels that carry no information about $A$. This supervision routing matches the physical reality: airlight is only observable where scattering is present.

\begin{figure}[H]
  \centering
  \includegraphics[width=\linewidth]{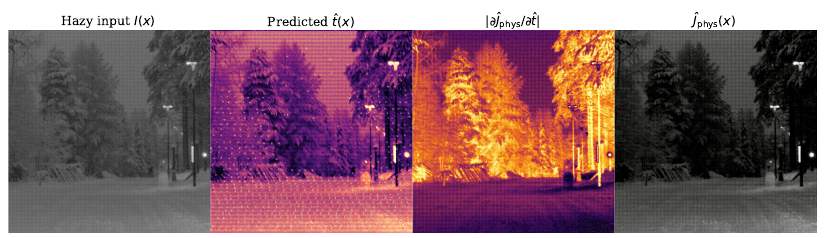}
  \caption{Empirical gradient map on a real STF fog frame. Left to right: hazy input, predicted transmission $\hat{t}$, analytical gradient magnitude map $|\partial\hat{J}_{\mathrm{phys}}/\partial\hat{t}|$, and physics-restored output. High-gradient regions align spatially with low-transmission regions, empirically validating the density-adaptive behaviour predicted by Propositions~1--2.}
  \label{fig:empirical_gradient_map}
\end{figure}

\paragraph{Empirical spatial validation.}
Figure~\ref{fig:empirical_gradient_map} confirms the $(\hat{t}+\epsilon)^{-2}$ scaling holds on real (non-synthetic) STF fog frames at the per-pixel level: the analytically predicted gradient magnitude map spatially aligns with low-transmission regions in the predicted $\hat{t}$, with high-gradient activations concentrated precisely where the physics model is most sensitive.

\paragraph{Contrast with the direct-head gradient.}
The direct head's gradient through the shared encoder, $\partial\hat{J}_{\mathrm{dir}}/\partial\mathbf{z} = \sigma'(\cdot)\,W_{\mathrm{dir}}$, is spatially uniform and has no notion of local fog density. The physics head provides spatially structured, density-adaptive magnitudes. Both signals flow into the same encoder, but the physics path shapes features toward physically meaningful representations even though only the direct head is used at inference.

\paragraph{Interpretive connection to inverse-variance weighting.}
Proposition~1 establishes that the stochastic sensitivity of the inversion scales as $t^{-2}$. Classical heteroscedastic regression~\cite{kendall2018multitasklearningusinguncertainty,mai2021batchinversevarianceweightingdeep,pmlr-v235-fisher24a} exploits this inversely, multiplying each sample's loss by $1/\operatorname{Var}[\hat{J}]\propto t^2$ to down-weight unreliable pixels. Differentiating through Eq.~\eqref{eq:phys_inv_eps_app} applies the structural inverse, up-weighting low-transmission regions. The two schemes are not in conflict; they operate on different signal models. Inverse-variance weighting is optimal when the signal is dominated by irreducible noise. Under the scattering model, the numerator $(\hat{A}-I(x)) = -(J-A)\,t + O(\xi)$ is dominated by the deterministic $(J-A)$ term in degraded regions where $(J-A)$ is large, so larger updates reduce estimation error rather than fit noise. We interpret the physics head as performing implicit hard-example mining along the transmission axis, empirically supported by the NoPhy ablation (§\ref{sec:physics_ablation}).

\paragraph{Bounded gradients and stability.}
To ensure numerical stability, the transmission is constrained to $\hat{t}\in[t_{\min}, t_{\max}]$ with $t_{\min}=0.05$ (Eq.~\ref{eq:head_trans}). This bounds the Jacobian:
\begin{equation}
    \left|\frac{\partial \hat{J}_{\mathrm{phys}}}{\partial \hat{t}}(x)\right|
    \leq \frac{|\hat{A}-I(x)|}{(t_{\min}+\epsilon)^2}
    \leq \frac{1}{(t_{\min}+\epsilon)^2},
\end{equation}
since $\hat{A}, I(x) \in [0,1]$. In practice, this upper bound is approximately $400$, and is further controlled via gradient clipping (Appendix~\ref{app:hyperparams}). As a result, the optimization remains within a bounded-gradient regime compatible with standard convergence guarantees for stochastic gradient methods~\cite{mai2021stabilityconvergencestochasticgradient,zhou2024convergenceadaptivegradientmethods}.

\paragraph{Implicit curriculum.}
The transmission-dependent gradient scaling induces an implicit curriculum \cite{DBLP:conf/icml/BengioLCW09} over training samples. Early in training, low-transmission (high-degradation) regions dominate the gradient signal due to their larger magnitude. As the model improves and transmission estimates stabilize, the gradient distribution naturally rebalances without requiring explicit scheduling. This behavior emerges directly from the physics-based formulation and does not introduce additional hyperparameters.

\paragraph{Scope.}
The analysis assumes (i) homogeneous single-scattering, (ii) Gaussian i.i.d.\ sensor noise, and (iii) first-order accuracy of $(\hat{t},\hat{A})$. Real fog may violate (i) through non-uniform airlight and multiple scattering; low-light imagery may violate (ii) through Poisson-dominated shot noise; heavily degraded regions may violate (iii) when $\delta\not\ll t_{\min}$. The structural amplification arises directly from the $1/\hat{t}$ factor in Eq.~\eqref{eq:phys_inv_eps_app}, not from Gaussianity—and is therefore robust under (ii); the prefactor may shift under (i) and (iii). The cross-domain results on O-HAZE and I-HAZE (§\ref{sec:ohaze}), where the NoPhy ablation degrades PSNR-Y by $1.53$ and $0.60$\,dB on real non-synthetic haze, provides empirical evidence that the mechanism transfers beyond the assumed regime.

\subsection{Loss Definitions}
\label{app:losses}
Let $\mathrm{Ch}(x,y)=\frac{1}{N}\sum\sqrt{(x-y)^2+\varepsilon^2}$ ($\varepsilon{=}10^{-6}$), $\partial_{x,y}$ finite differences, $\phi_l$ frozen VGG-16 features at relu1\_2, relu2\_2, relu3\_3, and $\psi$ VGG-16 up to relu2\_2.

\emph{Data-driven family (supervises the deployed output).}
$\mathcal{L}_{\text{rec}}=\mathrm{Ch}(\hat J_{\text{dir}},J)$;\;
$\mathcal{L}_{\text{perc}}=\sum_l\|\phi_l(\hat J_{\text{dir}})-\phi_l(J)\|_1$;\;
$\mathcal{L}_{\text{grad}}=\sum_{k\in\{x,y\}}\|\partial_k\hat J_{\text{dir}}-\partial_k J\|_1$.

\emph{Physics family (shapes optimization through the inversion).}
$\mathcal{L}_{\text{pr}}=\mathrm{Ch}(\hat J_{\text{phys}},J)$, the term carrying the $(\hat t+\epsilon)^{-2}$ amplification;
$\mathcal{L}_t=\|\hat t-t\|_1$ with $t=e^{-\beta\tilde d}$ from the synthesis depth, the anchor that localizes the amplified gradients;
$\mathcal{L}_{\text{sm}}=\sum_{k\in\{x,y\}}\overline{e^{-10|\partial_k I|}\,|\partial_k\hat t|}$, with $|\partial_k I|$ averaged over channels;
$\mathcal{L}_{\text{con}}=\mathrm{Ch}(\hat J_{\text{dir}},\hat J_{\text{phys}})$.

\emph{Auxiliary family.}
$\mathcal{L}_{\text{fft}}=\big\||\mathcal{F}(\hat J_{\text{dir}})|-|\mathcal{F}(J)|\big\|_1$ (orthonormal real 2-D FFT);
$\mathcal{L}_{\text{contra}}=\max\!\big(0,\|\psi(\hat J_{\text{dir}})-\psi(J)\|_1-\|\psi(\hat J_{\text{dir}})-\psi(I)\|_1+m\big)$, $m{=}0.3$, enabled from epoch 185.

The objective is $\mathcal{L}=\sum_i\lambda_i\mathcal{L}_i$ with the weights in Table~\ref{tab:loss_weights}, shared across S/M/L. NoPhy removes the entire physics family.

\section{Edge Deployment: Full Results}
\label{app:edge_full}

This appendix provides the full deployment measurements summarized in
\S\ref{sec:edge_deployment}: INT8 quantization quality on the
fixed-function NPU, qualitative comparison under quantization, and
the full embedded-GPU benchmark including CUDA FP32 reference.
numbers and onboard power profiles.

\subsection{INT8 Quantization Quality on the Fixed-Function NPU}
\label{app:edge_quant}

To verify that INT8 post-training quantization preserves restoration
quality, we study PhysDNet-M in detail. We use the vendor compiler's
default calibration pipeline with 64 real STF samples
($\beta{=}0.02$, $A{=}0.8$). \textbf{We do not apply
quantization-aware training, mixed-precision layer policies, custom
activation-range search, or any other vendor-specific optimization};
the reported gap is therefore representative of naive post-training
quantization rather than a tuned result. Evaluation covers 200
held-out samples spanning $\beta\in\{0.005,\ldots,0.10\}$
(Table~\ref{tab:quant_psnr}).

Within the training range ($\beta{\leq}0.04$), INT8 incurs only a
${\sim}1.5$\,dB average PSNR drop vs.\ FP32; at $\beta{=}0.01$ the
quantized model comes within 0.8\,dB of FP32. Out-of-distribution
degradation ($\beta{>}0.04$) follows the same graceful decline as
FP32, confirming that the physics-guided head is not destabilized by
aggressive quantization.

\begin{table}
  \centering
  \caption{%
    \textbf{PhysDNet-M INT8 quantization quality} on the fixed-function
    NPU. PSNR drop vs.\ FP32 is ${\sim}1.5$\,dB within the training
    $\beta$ range. $\pm$ values are 1$\sigma$ over $n$ samples.}
  \label{tab:quant_psnr}
  \small
  \begin{tabular}{@{}lccc@{}}
    \toprule
    $\beta$ & $n$ & INT8 PSNR (dB) & FP32 PSNR (dB) \\
    \midrule
    0.005 & 29 & 35.30$\pm$4.9 & 37.95$\pm$4.9 \\
    0.010 & 29 & 36.38$\pm$3.9 & 37.19$\pm$5.1 \\
    0.020 & 29 & 34.08$\pm$5.8 & 35.58$\pm$5.9 \\
    0.040 & 29 & 31.42$\pm$6.1 & 32.34$\pm$6.5 \\
    \midrule
    \multicolumn{2}{@{}l}{\emph{In-distribution avg}} & 34.30 & 35.77 \\
    \midrule
    0.060 & 28 & 22.78$\pm$7.9 & 25.91$\pm$8.4 \\
    0.080 & 28 & 20.07$\pm$6.1 & 20.02$\pm$6.4 \\
    0.100 & 28 & 14.31$\pm$5.2 & 16.18$\pm$6.0 \\
    \midrule
    \multicolumn{2}{@{}l}{\emph{Overall avg}} & 27.89 & 29.31 \\
    \bottomrule
  \end{tabular}
\end{table}

\subsection{Embedded-GPU Full Benchmark}
\label{app:edge_gpu}

Table~\ref{tab:jetson-benchmark-full} reports the full benchmark
under both CUDA FP32 and TensorRT FP16 execution providers on the
Jetson Orin Nano. The TensorRT FP16 path yields $3$--$5\times$ speedups
over CUDA FP32 across the board with no measurable PSNR degradation,
and is therefore the reference deployment configuration in the main
text.

\begin{table}
\centering
\caption{%
  \textbf{Full embedded-GPU benchmark}
  ($256{\times}256$, 50 warmup + 300 timed iterations, MAXN mode).
  Mean and median latency, FPS, and onboard rail power reported for
  both CUDA (FP32) and TensorRT FP16 providers.}
\label{tab:jetson-benchmark-full}
\resizebox{\linewidth}{!}{%
\begin{tabular}{@{}lrlrrrrr@{}}
\toprule
Model & Params & Provider & Mean (ms) & Median (ms) & FPS
  & $\overline{P}$ (W) & $P_{\max}$ (W) \\
\midrule
AOD-Net LiDAR        & 0.002\,M & TRT\_FP16 &   3.06 &   3.05 & 327.0 &  4.94 &  5.21 \\
AOD-Net LiDAR        & 0.002\,M & CUDA      &  12.36 &  12.35 &  80.9 &  5.13 &  5.66 \\
FFA-Net LiDAR        & 0.77\,M  & TRT\_FP16 &  21.31 &  21.30 &  46.9 &  8.21 & 10.09 \\
FFA-Net LiDAR        & 0.77\,M  & CUDA      &  83.67 &  83.65 &  12.0 & 13.39 & 15.44 \\
DehazeFormer-T LiDAR & 1.36\,M  & TRT\_FP16 &  21.25 &  21.24 &  47.1 & 12.38 & 12.93 \\
DehazeFormer-T LiDAR & 1.36\,M  & CUDA      &  62.61 &  62.60 &  16.0 & 11.47 & 13.17 \\
DEA-Net LiDAR        & 3.65\,M  & TRT\_FP16 &  35.12 &  35.12 &  28.5 & 10.17 & 12.42 \\
DEA-Net LiDAR        & 3.65\,M  & CUDA      &  89.93 &  89.93 &  11.1 & 12.45 & 13.05 \\
\midrule
PhysDNet-S (ours)    & 0.72\,M  & TRT\_FP16 &   9.24 &   9.23 & 108.3 & 13.45 & 13.48 \\
PhysDNet-S (ours)    & 0.72\,M  & CUDA      &  40.72 &  40.70 &  24.6 & 10.63 & 13.54 \\
PhysDNet-M (ours)    & 2.95\,M  & TRT\_FP16 &  17.09 &  17.09 &  58.5 & 12.38 & 12.42 \\
PhysDNet-M (ours)    & 2.95\,M  & CUDA      &  79.08 &  79.07 &  12.6 & 12.78 & 14.97 \\
PhysDNet-L (ours)    & 6.59\,M  & TRT\_FP16 &  25.80 &  25.80 &  38.8 & 11.80 & 13.01 \\
PhysDNet-L (ours)    & 6.59\,M  & CUDA      & 120.52 & 120.54 &   8.3 & 13.33 & 16.12 \\
\bottomrule
\end{tabular}}
\end{table}

\paragraph{Hardware disclosure.}
The NVIDIA Jetson Orin Nano Super (8\,GB, MAXN) is named explicitly. The fixed-function NPU is characterised by its specifications (26\,TOPS INT8, $<$4\,W TDP, vendor dataflow compiler with default post-training quantization); all throughput and quality figures are as measured.

\section{Baseline Descriptions}
\label{app:baselines}
 
We compare PhysDNet against both established dehazing architectures
and architectural ablations to quantify the contribution of each
design choice.
 
\paragraph{Published baselines.}
\textbf{AOD-Net}~\cite{DBLP:conf/iccv/LiPWXF17}: a lightweight
all-in-one CNN that directly regresses the clean image, representing
early physics-reformulated dehazing.
\textbf{FFA-Net}~\cite{DBLP:conf/aaai/QinWBXJ20}: a feature-fusion
attention CNN; strong attention-based baseline.
\textbf{DehazeFormer-T}~\cite{Song_2023}: a Transformer-based
restoration model representing modern attention architectures.
\textbf{DEA-Net}~\cite{DBLP:journals/tip/ChenHL24}: a difference-aware encoder-decoder network. We include it to benchmark our physics-guided feature extraction against, purely data-driven detail enhancement.

\section{Model Efficiency Breakdown}
\label{app:flops}
 
Table~\ref{tab:flops} provides a detailed efficiency comparison
across all models.  PhysDNet variants are measured at $256{\times}256$;
baselines at $512{\times}512$ (their best-performing resolution per
Table~\ref{tab:main_comparison}).
 
\begin{table}
\centering
\caption{%
  \textbf{Full efficiency breakdown.}
  Latency and FPS on a single NVIDIA GPU, batch size 1.
  $^\dagger$Measured at $256{\times}256$. We report the PyTorch numbers (no additional optimization is conducted)
}
\label{tab:flops}
\begin{tabular}{llccccc}
\toprule
\textbf{Model} & \textbf{Config}
  & \textbf{Params\,(M)} & \textbf{GMACs} & \textbf{GFLOPs}
  & \textbf{Lat.\,(ms)} & \textbf{FPS} \\
\midrule
AOD-Net         & NIR            & 0.002 &   0.23 &   0.46 & 0.3 & 3\,753 \\
AOD-Net         & LiDAR          & 0.002 &   0.24 &   0.47 & 0.3 & 3\,550 \\
DehazeFormer-T  & NIR            & 1.36  &  10.92 &  21.83 & 7.0 &   143  \\
DehazeFormer-T  & LiDAR          & 1.36  &  10.97 &  21.95 & 7.0 &   143  \\
FFA-Net         & NIR            & 0.77  &  98.57 & 197.15 & 11.5 &   87  \\
FFA-Net         & LiDAR          & 0.77  &  98.73 & 197.45 & 11.5 &   87  \\
DEA-Net         & RGB            & 3.65  &  16.42 &  32.84 & 8.9 & 112.6 \\
DEA-Net         & LiDAR          & 3.65  &  16.44 &  32.88 & 8.9 & 112.6 \\

\midrule
PhysDNet-S$^\dagger$ & ch32 & 0.72 &  2.24 &  4.47 & 1.5 &  661 \\
PhysDNet-M$^\dagger$ & ch64, SE    & 2.95 &  8.56 & 17.13 & 2.3 &  429 \\
PhysDNet-L$^\dagger$ & ch96, SE    & 6.59 & 18.98 & 37.97 & 3.6 &  275 \\
\bottomrule
\end{tabular}%
\end{table}
\section{Auxiliary Loss Ablation}
\label{app:ablation}

We ablate the two auxiliary loss components, FFT and Contrastive, that supplement the base objective (reconstruction + perceptual + gradient). We trained PhysDNet-M (ch64) for 200 epochs at $512{\times}512$; each row in Table~\ref{tab:ablation} enables one auxiliary loss on top of the base objective. The results motivate the inclusion of both in the complete recipe (\cref{tab:loss_weights}), which combines them under the extended 500-epoch schedule used in the main results.

\begin{table}
\centering
\caption{%
  \textbf{Auxiliary loss ablation.} PhysDNet-M (ch64) trained for 200 epochs at $512{\times}512$. Each row enables one auxiliary loss on top of the base (reconstruction + perceptual + gradient) objective. FFT contributes the largest PSNR gain; Contrastive contributes the largest SSIM gain.
}
\label{tab:ablation}
\begin{tabular}{lcc}
\toprule
\textbf{Auxiliary loss} & \textbf{PSNR (dB)\,$\uparrow$} & \textbf{SSIM\,$\uparrow$} \\
\midrule
Base objective only     & 32.38 & 0.9315 \\
\quad + Contrastive     & 32.97 & \textbf{0.9406} \\
\quad + FFT             & \textbf{33.02} & 0.9401 \\
\bottomrule
\end{tabular}
\vspace{-2em}
\end{table}
 
\section{Downstream Detection Protocol}
\label{app:detection_protocol}
 
\textbf{Fog generation.}
Synthetic fog is applied using the same Koschmieder model as in
training (\S\ref{sec:setup}), with $\beta{=}0.06$ and $A{=}0.8$
to test at out-of-distribution severity.
 
\textbf{Detector and classes.}
YOLOv8-M~\cite{yaseen2024yolov8indepthexplorationinternal}
(COCO-pretrained, conf$\geq$0.25, IoU$\geq$0.5).
STF KITTI-format labels are mapped to COCO classes and grouped
into three families: \emph{vehicle} (car, bus, truck),
\emph{person}, and \emph{cyclist}.
AP per family uses 101-point interpolation (COCO style).
 
\textbf{Tiled inference.}
Dehazing at $256{\times}256$ with overlapping tiles (stride 192),
linearly blended in the overlap to avoid boundary artifacts.
\section{Cross-Domain Generalization: Additional Details}\label{app:crossdomain}
 
\paragraph{Training-free alternative: chrominance transfer.}
The restored output on O-HAZE is structurally strong but
desaturated.  We use the model's predicted transmission $\hat{t}$
for \emph{transmission-guided chrominance transfer}: convert both
restored and hazy images to YCbCr, keep the Y channel from the
restored output, and transplant attenuated Cb/Cr from the hazy
input:
\[
  \mathrm{Cb}_{\mathrm{out}} = 0.5 + \hat{t} \cdot
    (\mathrm{Cb}_{\mathrm{hazy}} - 0.5), \quad
  \mathrm{Cr}_{\mathrm{out}} = 0.5 + \hat{t} \cdot
    (\mathrm{Cr}_{\mathrm{hazy}} - 0.5).
\]
This attenuates haze-coloured chrominance in heavily fogged regions
(low $\hat{t}$) while preserving natural colour where the scene is
clear (high $\hat{t}$).
Because it copies chrominance from the hazy input, the haze's colour cast persists; we therefore learn colour instead (App.~\ref{app:color_adapt}).
 
\begin{figure}[H]
\centering
\includegraphics[width=\linewidth]{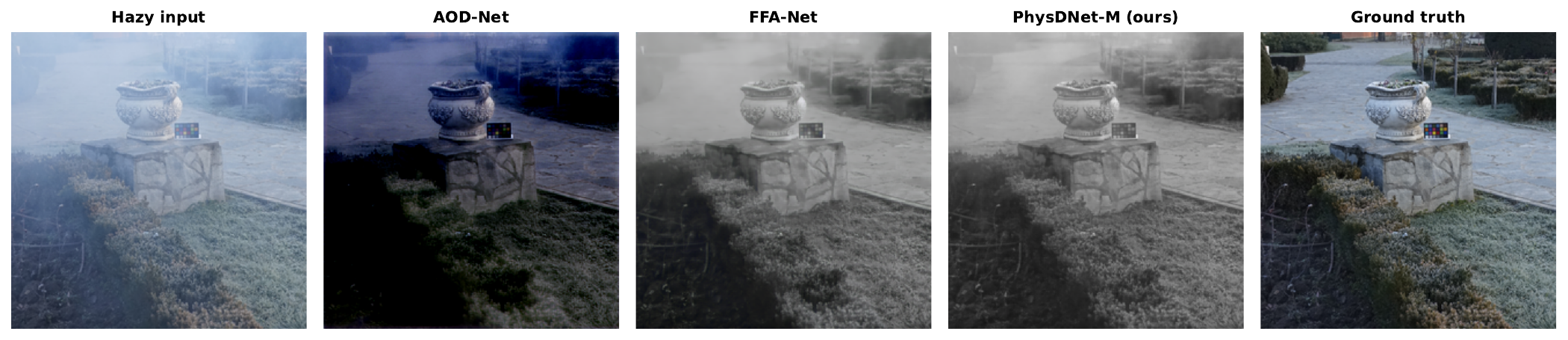}
\caption{%
  \textbf{Zero-shot O-HAZE restoration.}
  Left to right: hazy input, AOD-Net, FFA-Net, PhysDNet-M (ours),
  ground truth.  All models trained on STF only.
}
\label{fig:ohaze_qual}
\end{figure}
 
\begin{figure}[H]
\centering
\includegraphics[width=\linewidth]{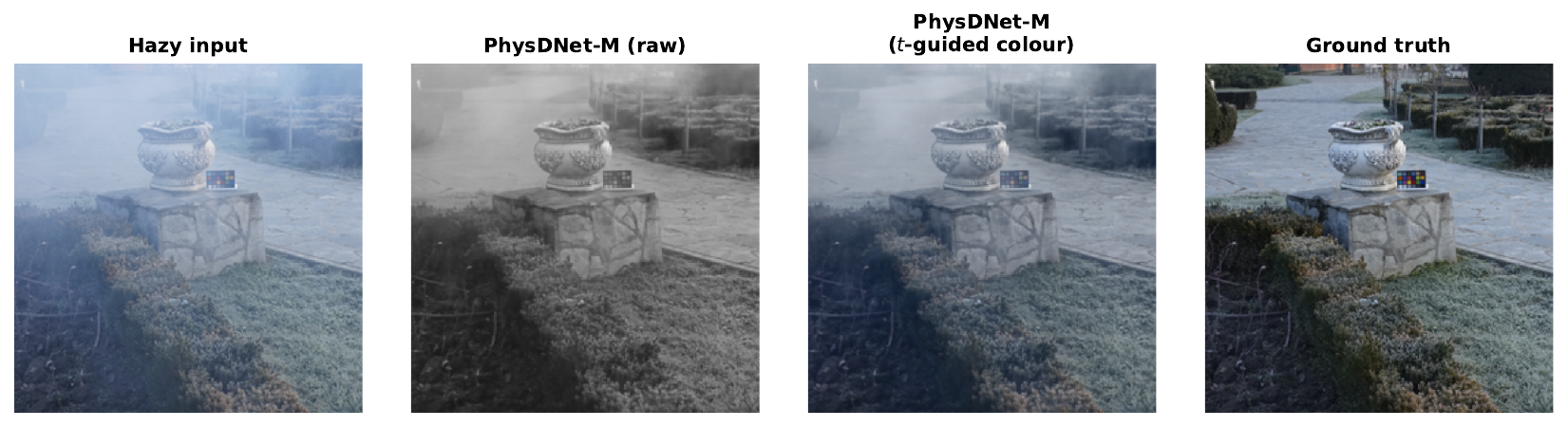}
\caption{%
  \textbf{Transmission-guided chrominance recovery.}
  Left to right: hazy input, raw PhysDNet-M output, after
  $\hat{t}$-guided chrominance transfer, ground truth.
}
\label{fig:ohaze_colour}
\end{figure}

\section{Self-Supervised Colour Adaptation and RTTS}
\label{app:color_adapt}
\paragraph{Objective.} On unpaired URHI images (4\,803, 5 epochs, encoders at reduced learning rate), PhysDNet-M is trained with:
(i) recomposition $\mathrm{Ch}(\hat J_{\text{dir}}\hat t+\hat A(1-\hat t),\,I)$;
(ii) dual-head consistency $\mathrm{Ch}(\hat J_{\text{dir}},\mathrm{sg}[\hat J_{\text{phys}}])$, where the stop-gradient makes the coloured inversion a teacher;
(iii) a dark-channel prior on $\hat J_{\text{dir}}$;
(iv) edge-aware $\hat t$ smoothness;
(v) supervised synthetic STF replay batches.
No paired data, labels, or architecture changes are used.

\section{Cross-Dataset Generalization: NH-HAZE}
\label{app:nhhaze}

NH-HAZE~\citep{Ancuti2020NHHAZEAI} contains 55 \emph{non-homogeneous} 
real outdoor hazy/clean image pairs (NTIRE 2020). Unlike O-HAZE's 
uniform fog, NH-HAZE exhibits spatially varying haze density, 
directly testing whether the physics head transfers despite the 
Koschmieder model's homogeneous-scattering assumption. As with other 
cross-dataset benchmarks, we feed $\mathbf{d}{=}\mathbf{0}$, $\mathbf{m}{=}\mathbf{0}$ and resize to $256{\times}256$. No 
fine-tuning.

\begin{table}[h]
\centering
\caption{%
  \textbf{Zero-shot NH-HAZE results at $256{\times}256$.}
  All models trained on STF; no NH-HAZE fine-tuning. Best in 
  \textbf{bold}.}
\label{tab:nhhaze}
\smallskip
\small
\begin{tabular}{@{}lcccc@{}}
\toprule
Method & PSNR-Y\,$\uparrow$ & SSIM-Y\,$\uparrow$ 
       & PSNR-RGB\,$\uparrow$ & SSIM-RGB\,$\uparrow$ \\
\midrule
AOD-Net         & 12.21 & 0.3835 & 12.12 & 0.3791 \\
FFA-Net         & 14.54 & 0.5037 & 14.16 & 0.4889 \\
DehazeFormer-T  & 14.55 & 0.4889 & 14.18 & 0.4742 \\
DEA-Net         & 14.59 & 0.5069 & 14.24 & 0.4932 \\
\midrule
PhysDNet-M (NoPhy)   & 14.20 & 0.5294 & 13.98 & 0.5165 \\
PhysDNet-S (ours)    & 14.70 & 0.4950 & 14.37 & 0.4807 \\
PhysDNet-M (ours)    & \textbf{14.74} & 0.5111 & \textbf{14.37} & 0.4965 \\
PhysDNet-L (ours)    & 14.45 & \textbf{0.5159} & 14.09 & \textbf{0.5012} \\
\bottomrule
\end{tabular}
\end{table}

PhysDNet-M achieves the highest PSNR-Y (14.74\,dB), outperforming 
DEA-Net by $+0.15$\,dB and FFA-Net by $+0.20$\,dB; PhysDNet-L 
attains the best SSIM-Y (0.5159). PhysDNet-S remains competitive 
(14.70\,dB) at $4{\times}$ fewer parameters than the M variant. 
Removing the physics head (NoPhy) reduces PSNR-Y by $0.54$\,dB and 
drops the model below all modern baselines, confirming that the 
physics-head contribution transfers to non-homogeneous fog despite 
the underlying Koschmieder model assuming homogeneous scattering. 
This evidence directly addresses the model-mismatch concern raised 
in our Limitations.

\section{LiDAR Sparsity Robustness}
\label{app:lidar_sparsity}

Automotive LiDAR produces inherently sparse depth maps---typically
$<$5\% of pixels carry valid returns, with sparsity varying
significantly across weather conditions, surface reflectivity, and
sensor occlusion. To study how PhysDNet's restoration quality
degrades as depth information becomes scarcer, we systematically
subsample valid LiDAR returns from 100\% down to 0\% (pure NIR
fallback) while keeping all other inputs fixed. This stress-tests
the gated-fusion mechanism (\S\ref{sec:architecture}): a well-designed
gate should learn to fall back to NIR features when depth is absent
($\mathbf{m}{=}\mathbf{0}$), rather than catastrophically failing.

\begin{figure}
  \centering
  \includegraphics[width=\linewidth]{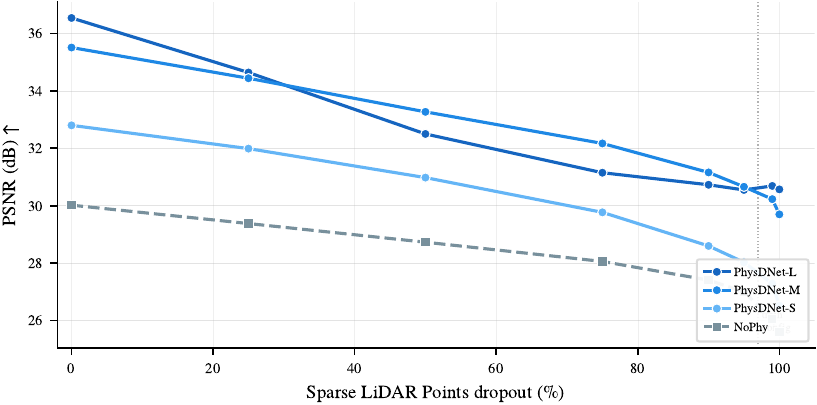}
  \caption{%
    \textbf{PSNR vs.\ LiDAR keep ratio.} All PhysDNet variants
    degrade gracefully as LiDAR points are removed. Even at 0\%
    density (NIR only), PhysDNet-M retains a $+4.10$\,dB advantage
    over the NoPhy ablation, confirming that the physics-guided
    representation transfers across input modalities. Vertical
    dotted line marks the typical training-time density
    ($\sim$3\%).}
  \label{fig:lidar_sparsity}
\end{figure}

\paragraph{Protocol.}
For each keep ratio $r \in \{1.0, 0.75, 0.50, 0.25, 0.10, 0.05, 0.01,
0.0\}$, we randomly retain $r$ fraction of valid mask pixels in the
STF validation set (1\,299 images) and zero-out the corresponding
depth entries as shown in \cref{tab:lidar_sparsity} and \cref{fig:lidar_sparsity}. The model receives a hazy image $I \in [0,1]^{3 \times H \times W}$, sparse LiDAR depth $D_s$, and validity mask $M$, and jointly predicts a restored image, transmission map, and airlight vector.

\begin{table}
\centering
\caption{%
  \textbf{LiDAR sparsity ablation} on STF val ($256{\times}256$).
  PSNR (dB) at each keep ratio. At $r{=}0$ (no LiDAR) the model falls
  back to NIR-only restoration via the learned gated-fusion mechanism.
  Training uses $r{\approx}3\%$ (typical Velodyne HDL-64 density).}
\label{tab:lidar_sparsity}
\small
\begin{tabular}{@{}lcccccccc@{}}
\toprule
\textbf{Model} & \textbf{100\%} & \textbf{75\%} & \textbf{50\%}
  & \textbf{25\%} & \textbf{10\%} & \textbf{5\%} & \textbf{1\%}
  & \textbf{0\%} \\
\midrule
PhysDNet-S         & 32.80 & 31.99 & 30.98 & 29.77 & 28.60 & 28.03 & 27.29 & 26.52 \\
PhysDNet-M         & 35.51 & 34.44 & 33.27 & 32.17 & 31.16 & 30.66 & 30.23 & 29.70 \\
PhysDNet-L         & 36.54 & 34.64 & 32.50 & 31.15 & 30.73 & 30.55 & 30.69 & 30.57 \\
\midrule
PhysDNet-M (NoPhy) & 30.02 & 29.38 & 28.73 & 28.06 & 27.41 & 27.00 & 26.06 & 25.60 \\
\bottomrule
\end{tabular}
\end{table}

\paragraph{Three observations.}
\textbf{(i)}~All PhysDNet variants degrade smoothly rather than
collapsing as LiDAR density decreases, confirming the gated-fusion
mechanism (\S\ref{sec:architecture}) successfully learned to attribute
weight to NIR features when depth signal is unavailable.
\textbf{(ii)}~Even at 0\% LiDAR (pure NIR fallback), PhysDNet-M
retains $29.70$\,dB---still $4.10$\,dB above the NoPhy ablation at
matched 0\% density ($25.60$\,dB). This confirms that the physics
head's contribution is not merely depth-driven but reshapes the
encoder's NIR processing as well.
\textbf{(iii)}~At extreme sparsity ($r{=}1\%$, well below training
density), PhysDNet-L's PSNR ($30.69$) remains higher than at $r{=}0\%$
($30.57$), indicating that even one valid LiDAR point per 100 pixels
provides usable signal. This degradation profile validates deployment
under realistic LiDAR conditions where weather can drop point density
significantly below nominal.

\section{No-Reference Perceptual Quality}
\label{app:niqe}

Models trained on STF synthetic fog produce near-achromatic
restorations on cross-dataset images (\S\ref{sec:ohaze}), so
chrominance-sensitive full-reference metrics underestimate
restoration quality. We supplement Table~\ref{tab:crossdomain}
with two established no-reference IQA metrics:
\textbf{NIQE}~\cite{mittal2013making} and
\textbf{BRISQUE}~\cite{Mittal} (lower $=$ better; both assess
naturalness and distortion blindly without ground truth). We
evaluate on O-HAZE (45 outdoor pairs, no LiDAR;
$\mathbf{d}{=}\mathbf{0}$).

\begin{table}[H]
\centering
\caption{%
  \textbf{No-reference perceptual quality on O-HAZE}
  (NIQE\,/\,BRISQUE, lower $=$ better). Numbers in parentheses are
  1$\sigma$ across the 45-image O-HAZE validation set. ``Hazy'' is
  the unprocessed input. Best in \textbf{bold}.}
\label{tab:niqe}
\small
\begin{tabular}{@{}lcccc@{}}
\toprule
\textbf{Method}
  & \multicolumn{2}{c}{\textbf{NIQE\,$\downarrow$}}
  & \multicolumn{2}{c}{\textbf{BRISQUE\,$\downarrow$}} \\
\midrule
Hazy input          &  4.33 & ($\sigma$=1.14) &  7.75 & ($\sigma$=7.02) \\
\midrule
AOD-Net             &  6.37 & ($\sigma$=4.18) & 20.62 & ($\sigma$=15.58) \\
FFA-Net             &  4.30 & ($\sigma$=1.41) & 12.25 & ($\sigma$=8.84) \\
DehazeFormer-T      &  4.32 & ($\sigma$=1.25) & 12.76 & ($\sigma$=9.83) \\
DEA-Net             &  4.64 & ($\sigma$=1.62) & 11.95 & ($\sigma$=10.66) \\
\midrule
PhysDNet-S          &  4.21 & ($\sigma$=1.04) & 15.98 & ($\sigma$=6.92) \\
PhysDNet-M          &  \textbf{3.92} & ($\sigma$=1.08) & 11.13 & ($\sigma$=5.58) \\
PhysDNet-L          &  4.06 & ($\sigma$=1.10) & \textbf{9.49} & ($\sigma$=6.73) \\
PhysDNet-M (NoPhy)  &  3.91 & ($\sigma$=1.27) & 11.96 & ($\sigma$=6.29) \\
\bottomrule
\end{tabular}
\end{table}

\paragraph{Discussion.}
PhysDNet leads or matches the strongest baselines on both
no-reference metrics. PhysDNet-M achieves the best NIQE among
learned methods ($3.92$, vs.\ DEA-Net $4.64$ and AOD-Net $6.37$);
PhysDNet-L achieves the best BRISQUE ($9.49$, vs.\ FFA-Net $12.25$
and DEA-Net $11.95$). Compared to the NoPhy ablation, PhysDNet-M
shows a consistent BRISQUE improvement ($-0.83$), while NIQE
differences fall within noise---NIQE is a weaker discriminator than
BRISQUE for low-distortion images, where most methods already
approach naturalness statistics. Together with the PSNR-Y and
SSIM-Y results in Table~\ref{tab:crossdomain}, these no-reference
metrics provide independent confirmation that PhysDNet's
restoration quality on real haze is not an artifact of
synthetic-data circularity.

\section{Adverse-Weather Consistency and Failure Modes}
\label{app:weather}
\begin{table}[H]\centering\small
\caption{Dual-head agreement (PSNR / Pearson $r$) on real STF captures, 250 frames per official split, full resolution, released PhysDNet-M checkpoint.}
\label{tab:weather}
\begin{tabular}{@{}lcc@{}}
\toprule
Condition & Day & Night \\
\midrule
Clear     & 39.15 / 0.998 & 33.39 / 0.993 \\
Light fog & 40.13 / 0.999 & 34.73 / 0.994 \\
Dense fog & \textbf{40.31} / 0.999 & 34.84 / 0.995 \\
Snow      & 39.45 / 0.997 & 35.38 / 0.994 \\
Rain      & 35.05 / 0.995 & --- \\
\bottomrule
\end{tabular}
\end{table}
Agreement is a consistency measure, not accuracy against clean references.

\paragraph{Failure modes and restrictions.}
\textbf{Night:} a uniform ${\approx}5$\,dB consistency reduction ($r{\geq}0.993$); active light sources locally violate the global-airlight assumption, and the Direct Head absorbs this.
\textbf{Over-enhancement:} $\hat t{\geq}t_{\min}{=}0.05$ caps the inversion gain at ${\approx}20\times$ by design.
\textbf{Colour:} after adaptation, colour inherits $\hat A$, so a mild warm shift can appear under tinted illumination.
\textbf{Adaptation trade-off:} $-5.1$\,dB on O-HAZE.
\textbf{Low-transmission noise:} in the Physics Head this is the $\sigma^2/t^2$ structure of Prop.~1 (App.~\ref{app:prop1_validation}), not inherited by the deployed Direct Head.

\section{RGB-Only Training and Other Inverse Problems}
\label{app:rgb_scope}
\paragraph{RESIDE-6K.} PhysDNet-M and a matched NoPhy twin are trained from scratch (RGB only, $D_s{=}M{=}0$, $\lambda_t{=}0$, same seed and recipe, 600 epochs). On the full 1\,000-image SOTS-mix test set they reach 28.45 vs.\ 28.59\,dB, and 21.9 vs.\ 22.2\,dB on the densest-pixel bin. Depth-branch folding is bit-exact (max $|\Delta|{=}0$) at 6.33 vs.\ 8.56\,GMACs. Jetson TensorRT FP16 latency is unchanged (13.3\,ms, bandwidth-bound), so the saving accrues on MAC-limited NPUs. Zero-shot on O-HAZE, the STF-trained model exceeds this colour-native RESIDE model by $+3.1$\,dB PSNR-RGB (19.65 vs.\ 16.58).

\paragraph{Underwater restoration (anchored).} From NYU-v2 RGB-D~\cite{silberman2012nyu} we synthesise $I_c=J_c e^{-\beta_c d}+B_c(1-e^{-\beta_c d})$~\cite{akkaynak2018underwater} with $\beta_r\gg\beta_g>\beta_b$ and signal-dependent noise. A 3\% sparse-depth prior anchors a 3-channel transmission head, mirroring LiDAR on STF. Comparison axes were fixed before the runs.
\begin{table}[H]\centering\small
\caption{Underwater (NYU-v2 synthesis, 1\,000 epochs, seed 42). PSNR (dB) by ground-truth transmission bin.}
\label{tab:underwater}
\begin{tabular}{@{}lccccc@{}}
\toprule
 & Overall & Dense ($t{<}0.2$) & Mid & Light & Red ch. \\
\midrule
Physics & \textbf{28.60} & \textbf{24.04} & \textbf{31.58} & 37.30 & \textbf{24.28} \\
NoPhy   & 28.22 & 23.83 & 30.76 & \textbf{37.37} & 23.68 \\
\bottomrule
\end{tabular}
\end{table}

\paragraph{Low-light enhancement.} The Retinex inversion $\hat R=I/(\hat L+\epsilon)$ is the $A{\to}0$ case of Eq.~\eqref{eq:physics_inversion}. On LOL-v2-Real~\cite{yang2021lolv2} (100 test images), the physics variant leads in both run pairings (20.04/19.64 vs.\ 19.46/19.15\,dB; SSIM 0.906 vs.\ 0.897). On LOL-v1~\cite{wei2018retinex}, the 15-image test set is noise-dominated and the sign is unstable, so the result is inconclusive.

\paragraph{Summary.} Amplification helps where the inverted parameter is anchored by a measurement and degradation is dense, and is neutral otherwise. Magnitudes outside STF (0.2--0.8\,dB) come from a single seed.

\section{Dataset Details}
\label{app:stf_details}

\begin{figure}
\centering
\includegraphics[width=\textwidth]{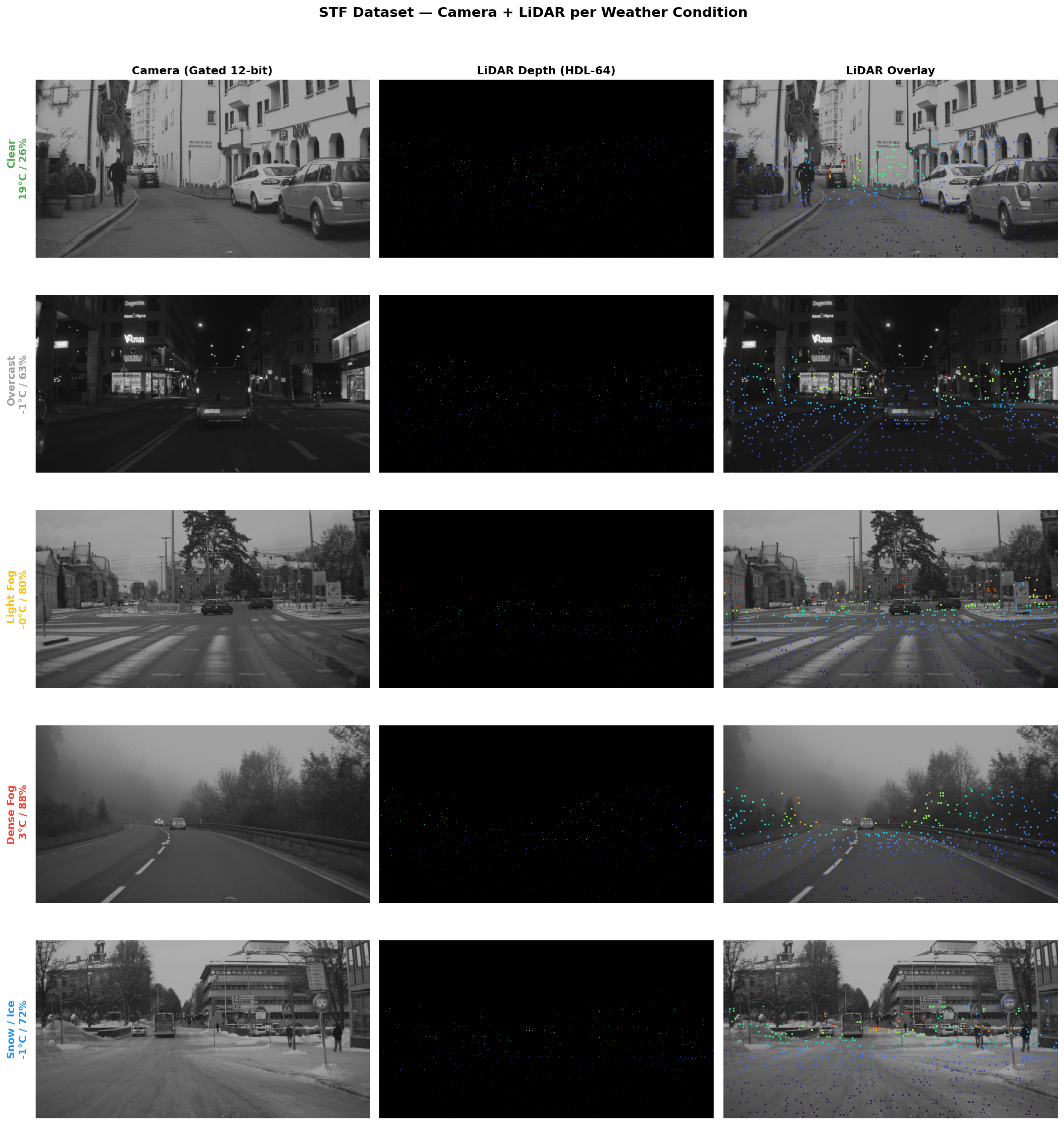}
\caption{%
  \textbf{STF sensor overlay per weather condition.}
  Sparse Velodyne HDL-64 LiDAR returns (centre, right) highlight
  the multi-modal perception challenge addressed by PhysDNet.
  Imagery from the monochrome 12-bit gated NIR camera.
}
\label{fig:app_stf_sensor_overlay}
\end{figure}

The Seeing Through Fog (STF)
benchmark~\cite{DBLP:conf/cvpr/BijelicGMKRDH20} provides 10\,280
synchronised multi-sensor frames captured over 14 months of real
adverse weather across Northern and Central Europe, spanning a
temperature range of $-25.5^{\circ}$C to $19.7^{\circ}$C. Each frame
includes monochrome 12-bit gated NIR imagery, sparse Velodyne HDL-64
LiDAR returns, RGB stereo, and weather metadata. We use only the
gated NIR and LiDAR modalities, as they are the most informative under
fog (\S\ref{app:ext_background}). Table~\ref{tab:stf_summary_stats}
summarises the per-condition split and how each subset is used in
our experiments. Figure~\ref{fig:app_stf_sensor_overlay} illustrates
the multi-modal perception challenge: LiDAR returns are sparse and
non-uniform across weather conditions, motivating our gated-fusion
design (\S\ref{sec:architecture}).

\begin{table}
\centering
\caption{%
  \textbf{STF dataset split statistics.}
  10\,280 frames with weather metadata, Feb 2018--May 2019,
  $-25.5^{\circ}$C to $19.7^{\circ}$C.
}
\label{tab:stf_summary_stats}
\begin{tabular}{lrrc}
\toprule
\textbf{Condition} & \textbf{Frames} & \textbf{Ratio}
  & \textbf{Usage} \\
\midrule
Clear     & 1\,497 & 14.6\% & Synthetic haze training \\
Overcast  & 2\,715 & 26.4\% & Synthetic haze training \\
Light Fog & 3\,788 & 36.8\% & Real-fog eval (Table~\ref{tab:consistency}) \\
Dense Fog & 1\,219 & 11.9\% & Real-fog eval (Table~\ref{tab:consistency}) \\
Snow/Ice  & 1\,061 & 10.3\% & Consistency eval (App.~\ref{app:weather}) \\
\midrule
\textbf{Total} & \textbf{10\,280} & \textbf{100\%} & \\
\bottomrule
\end{tabular}
\vspace{-1.5em}
\end{table}

\section{Additional Qualitative Results}
\label{app:qualitative}

\begin{figure}
\centering
\begin{minipage}{\linewidth}
\centering
\small
\begin{tabular}{@{}p{0.167\linewidth}@{}p{0.167\linewidth}@{}p{0.167\linewidth}@{}p{0.167\linewidth}@{}p{0.167\linewidth}@{}p{0.167\linewidth}@{}}
\centering Hazy Input &
\centering Direct Head &
\centering Physics Head &
\centering Ground Truth &
\centering Transmission &
\centering\arraybackslash LiDAR Depth \\
\end{tabular}
\end{minipage}\\[-2pt]
\includegraphics[width=\linewidth]{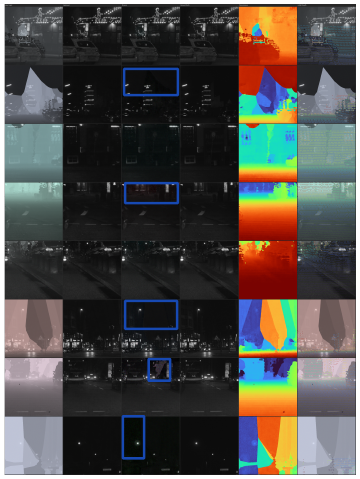}
\caption{%
  \textbf{Dual-head outputs on synthetic STF fog.}
  Hazy inputs are rendered with Eq.~\eqref{eq:koschmieder} from nearest-neighbour-filled sparse LiDAR depth, so haze density follows the interpolated depth and can look unnatural at depth discontinuities and in regions without LiDAR returns (e.g.\ sky). Blue boxes mark dense-haze regions where the fixed, parameter-free Physics Head retains these projection artifacts and the $\sigma^2/t^2$ noise predicted by Prop.~1; the deployed Direct Head suppresses them, restoring beyond the Koschmieder model.}
\label{fig:qualitative_results}
\end{figure}

We provide additional dual-head visualisations to illustrate the
complementarity discussed in \S\ref{sec:heads}. Figure~\ref{fig:qualitative_results}
shows side-by-side outputs of the direct head and the physics head across eight diverse synthetic-fog samples from the STF validation set, alongside the predicted transmission map and sparse LiDAR overlay. The physics head produces sharper outputs in well-conditioned regions but exhibits noise amplification under dense haze (blue boxes in \cref{fig:qualitative_results}), where the $(\hat{t}+\epsilon)^{-2}$ inversion is most
ill-conditioned. The unconstrained direct head, trained jointly under the physics-anchored objective (\S\ref{sec:loss}), successfully suppresses these artifacts while inheriting the density-adaptive representation shaped during training. Only the direct head is deployed at inference.
 
\section{Out-of-Distribution Fog Robustness}
\label{app:beta_sweep}

\begin{figure}[H]
  \centering
  \includegraphics[width=\linewidth]{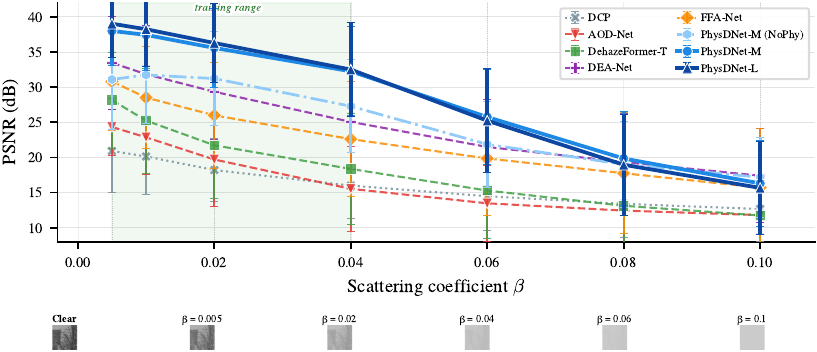}
  \caption{%
    \textbf{PSNR vs.\ fog density $\beta$.}
    All models evaluated at $256{\times}256$ with $A{=}0.8$.
    The green band denotes the training range.
    PhysDNet-M maintains the highest PSNR across all densities,
    retaining a ${>}$6\,dB advantage over the best baseline at
    $\beta{=}0.06$ and degrading gracefully outside the training
    distribution.
  }
  \label{fig:beta_sweep}
\end{figure}

To evaluate model robustness to out-of-distribution weather severities, we sweep the scattering coefficient $\beta \in [0.005, 0.10]$ at fixed global airlight ($A=0.8$) across the full 1,299-image validation set. As shown in \cref{fig:beta_sweep}, values $\beta \geq 0.05$ fall entirely outside the training distribution.

PhysDNet-M degrades smoothly beyond the training boundary, whereas baselines like FFA-Net suffer steeper drop-offs. We attribute this to the physics-constrained encoder: because the inversion formula is domain-invariant, the learned representations generalize significantly better to unseen $\beta$ values.

\newpage
\section*{NeurIPS Paper Checklist}

\begin{enumerate}

\item {\bf Claims}
    \item[] Question: Do the main claims made in the abstract and introduction accurately reflect the paper's contributions and scope?
    \item[] Answer: \answerYes{} 
    \item[] Justification: The abstract and introduction state four claims: (i) a density-adaptive gradient amplification mechanism arising from differentiation through a Koschmieder inversion (\cref{sec:gradient}, \cref{sec:theory_proof}); (ii) state-of-the-art accuracy on Seeing Through Fog at low compute (\cref{tab:main_comparison}, \cref{fig:gmacs_vs_psnr}); (iii) zero-shot generalization to non-homogeneous fog (\cref{sec:ohaze}, \cref{app:nhhaze}); and (iv) deployability on fixed-function NPUs (\cref{tab:edge-merged}, \cref{fig:hailo_sample}). All are quantitatively supported in the corresponding sections and appendices.

\item {\bf Limitations}
    \item[] Question: Does the paper discuss the limitations of the work performed by the authors?
    \item[] Answer: \answerYes{} 
    \item[] Justification:  A dedicated Limitations paragraph at the end of the Conclusion discusses (i) synthetic Koschmieder training distribution, (ii) STF-only multi-modal training, (iii) near-achromatic outputs from monochrome NIR supervision, and (iv) Koschmieder-specific theoretical scope. \cref{sec:theory_proof} also includes an explicit Scope paragraph stating the assumptions of the theoretical analysis (homogeneous single-scattering, Gaussian i.i.d.\ noise, first-order accuracy of $(\hat{t},\hat{A})$).

\item {\bf Theory assumptions and proofs}
    \item[] Question: For each theoretical result, does the paper provide the full set of assumptions and a complete (and correct) proof?
    \item[] Answer: \answerYes{} 
    \item[] Justification: Theoretical results in \cref{sec:gradient} (\cref{eq:grad_t,eq:grad_A}) are derived in full in \cref{sec:theory_proof}, including Proposition~1 (transmission-scaled noise variance with proof), Proposition~2 (Jacobian structure with derivation), the loss-gradient corollary, the airlight-gradient analysis, and the bounded-gradient analysis. Assumptions are stated explicitly in the Scope paragraph.

    \item {\bf Experimental result reproducibility}
    \item[] Question: Does the paper fully disclose all the information needed to reproduce the main experimental results of the paper to the extent that it affects the main claims and/or conclusions of the paper (regardless of whether the code and data are provided or not)?
    \item[] Answer: \answerYes{} 
    \item[] Justification: Training recipe (\cref{tab:recipe}), loss weights and activation schedule (\cref{tab:loss_weights}), model variants (\cref{tab:variants}), architecture details (\cref{app:arch_details}), dataset splits and statistics (\cref{app:stf_details}, \cref{tab:stf_summary_stats}), evaluation protocols for each experiment (\cref{app:detection_protocol,app:edge_full,app:beta_sweep,app:lidar_sparsity,app:niqe}), and baseline configurations (\cref{app:baselines}) are documented. Code available under \url{https://github.com/Heatdh/PhysDnet}.

\item {\bf Open access to data and code}
    \item[] Question: Does the paper provide open access to the data and code, with sufficient instructions to faithfully reproduce the main experimental results, as described in supplemental material?
    \item[] Answer: \answerYes{}{} 
    \item[] Justification: Code is available under \url{https://github.com/Heatdh/PhysDnet}, as stated in the abstract. STF~\citep{DBLP:conf/cvpr/BijelicGMKRDH20}, O-HAZE~\citep{ancuti2018ohazedehazingbenchmarkreal}, I-HAZE~\citep{DBLP:conf/acivs/AncutiATV18}, and NH-HAZE~\citep{Ancuti2020NHHAZEAI} are publicly available from their original authors. Hyperparameters, training recipes, and architectural details are fully documented in \cref{app:hyperparams,app:arch_details} to enable independent reimplementation.

\item {\bf Experimental setting/details}
    \item[] Question: Does the paper specify all the training and test details (e.g., data splits, hyperparameters, how they were chosen, type of optimizer) necessary to understand the results?
    \item[] Answer: \answerYes{} 
    \item[] Justification: Optimizer (AdamW), learning rate ($2{\times}10^{-4}$), schedule (cosine annealing with 10-epoch warmup), batch size, weight decay, gradient clipping, crop size, and mixed-precision use are all specified in \cref{tab:recipe}. Loss weights and the contrastive activation schedule (epoch 185) are in \cref{tab:loss_weights}. Haze synthesis parameters ($\beta\sim\mathcal{U}(0.005,0.04)$, $A_k\sim\mathcal{U}(0.7,1.0)$) are specified in the same table. Train/val splits are documented in \cref{app:stf_details}.

\item {\bf Experiment statistical significance}
    \item[] Question: Does the paper report error bars suitably and correctly defined or other appropriate information about the statistical significance of the experiments?
    \item[] Answer: \answerYes{} 
    \item[] Justification: PhysDNet-M reports mean$\pm$SE over 3 seeds in the main comparison (\cref{tab:main_comparison}: $35.55\pm0.07$\,dB PSNR, $0.9566\pm0.0009$ SSIM). INT8 NPU quantization quality reports $\pm 1\sigma$ across 28-29 samples per fog density $\beta$ (\cref{tab:quant_psnr}). No-reference perceptual metrics on O-HAZE include $1\sigma$ across the 45-image validation set (\cref{tab:niqe}). The fog-density sweep (\cref{fig:beta_sweep})  report per-condition averages over the full 1,299-frame validation set. Cross-domain results (\cref{tab:crossdomain,tab:nhhaze}) and downstream detection (\cref{tab:downstream}) report point estimates due to compute constraints; the magnitude of the effects ($+1.67$\,dB on STF, $-5.51$\,dB NoPhy ablation, $+0.62$\,dB cross-domain) substantially exceeds typical seed variance.

\item {\bf Experiments compute resources}
    \item[] Question: For each experiment, does the paper provide sufficient information on the computer resources (type of compute workers, memory, time of execution) needed to reproduce the experiments?
    \item[] Answer: \answerYes{} 
    \item[] Justification: Training compute (500 epochs per variant, single GPU, AMP (FP16) enabled for the L variant) is documented in \cref{app:hyperparams}. Edge deployment compute is reported in detail: fixed-function NPU and Jetson Orin Nano latency, FPS, and onboard power in \cref{tab:edge-merged} and \cref{tab:jetson-benchmark-full}; desktop GPU latency in \cref{tab:flops}. Power profiles and warmup/timing protocols are specified in the captions of \cref{tab:edge-merged,tab:jetson-benchmark-full}.
    
\item {\bf Code of ethics}
    \item[] Question: Does the research conducted in the paper conform, in every respect, with the NeurIPS Code of Ethics \url{https://neurips.cc/public/EthicsGuidelines}?
    \item[] Answer: \answerYes{} 
    \item[] Justification: Our work uses publicly available datasets collected by their original authors with appropriate research-use licensing. No human-subjects research was conducted.


\item {\bf Broader impacts}
    \item[] Question: Does the paper discuss both potential positive societal impacts and negative societal impacts of the work performed?
    \item[] Answer: \answerYes{} 
    \item[] Justification:  \cref{app:broader_impact} discusses positive impacts (improved adverse-weather perception for autonomous driving and robotics, reduced weather-related accidents, Green AI benefits from sub-4\,W deployment) and negative impacts (dual-use potential for unconsented surveillance or military deployment, safety risks under out-of-distribution weather violating the Koschmieder assumption). Mitigation recommendations (redundant sensor fusion, uncertainty estimation, hardware fail-safes) are included.
    
\item {\bf Safeguards}
    \item[] Question: Does the paper describe safeguards that have been put in place for responsible release of data or models that have a high risk for misuse (e.g., pre-trained language models, image generators, or scraped datasets)?
    \item[] Answer: \answerNA{} 
    \item[] Justification: This work releases a dehazing model trained on publicly available adverse-weather datasets. It does not pose elevated misuse risks comparable to large-scale generative models or scraped datasets. Dual-use considerations and recommended deployment safeguards are discussed in \cref{app:broader_impact}.

\item {\bf Licenses for existing assets}
    \item[] Question: Are the creators or original owners of assets (e.g., code, data, models), used in the paper, properly credited and are the license and terms of use explicitly mentioned and properly respected?
    \item[] Answer: \answerYes{} 
    \item[] Justification:All datasets (STF~\citep{DBLP:conf/cvpr/BijelicGMKRDH20}, O-HAZE~\citep{ancuti2018ohazedehazingbenchmarkreal}, I-HAZE~\citep{DBLP:conf/acivs/AncutiATV18}, NH-HAZE~\citep{Ancuti2020NHHAZEAI}) are cited and used under their original research-use licenses. Baselines (AOD-Net~\citep{DBLP:conf/iccv/LiPWXF17}, FFA-Net~\citep{DBLP:conf/aaai/QinWBXJ20}, DehazeFormer-T~\citep{Song_2023}, DEA-Net~\citep{DBLP:journals/tip/ChenHL24}) and YOLOv8-M~\citep{yaseen2024yolov8indepthexplorationinternal} are cited and used under their respective open-source licenses. Our code is hosted under \url{https://github.com/Heatdh/PhysDnet}.

\item {\bf New assets}
    \item[] Question: Are new assets introduced in the paper well documented and is the documentation provided alongside the assets?
    \item[] Answer: \answerYes{} 
    \item[] Justification: The PhysDNet model and code release will include model architecture, training pipeline, evaluation scripts, configuration files, and pre-trained checkpoints with accompanying documentation. Hyperparameters, architectural specifications, and training schedules are fully documented in the paper (\cref{app:hyperparams,app:arch_details}).

\item {\bf Crowdsourcing and research with human subjects}
    \item[] Question: For crowdsourcing experiments and research with human subjects, does the paper include the full text of instructions given to participants and screenshots, if applicable, as well as details about compensation (if any)? 
    \item[] Answer: \answerNA{} 
    \item[] Justification: No crowdsourcing or human-subjects research was conducted. All evaluations use publicly available datasets.

\item {\bf Institutional review board (IRB) approvals or equivalent for research with human subjects}
    \item[] Question: Does the paper describe potential risks incurred by study participants, whether such risks were disclosed to the subjects, and whether Institutional Review Board (IRB) approvals (or an equivalent approval/review based on the requirements of your country or institution) were obtained?
    \item[] Answer: \answerNA{} 
    \item[] Justification: No human-subjects research was conducted.

\item {\bf Declaration of LLM usage}
    \item[] Question: Does the paper describe the usage of LLMs if it is an important, original, or non-standard component of the core methods in this research? Note that if the LLM is used only for writing, editing, or formatting purposes and does \emph{not} impact the core methodology, scientific rigor, or originality of the research, declaration is not required.
    \item[] Answer: \answerNA{} 
    \item[] Justification: No large language models are used as a core component of the proposed method.

\end{enumerate}

\end{document}